\documentclass[12pt]{article}

\usepackage{newtxtext,newtxmath}

\usepackage{graphicx}

\usepackage[letterpaper,margin=1in]{geometry}

\renewenvironment{abstract}
	{\quotation}
	{\endquotation}

\date{}

\makeatletter
\renewcommand{\fnum@figure}{\textbf{Figure \thefigure}}
\renewcommand{\fnum@table}{\textbf{Table \thetable}}
\makeatother

\usepackage{scicite}

\usepackage{url}

\def\scititle{
	\textcolor{black}{Therapist-Exoskeleton-Patient Interaction \\ for Gait Therapy}
}
\title{\bfseries \boldmath \scititle}

\author
{Emek Barış Küçüktabak$^{1,2,4,\ast,\dagger,\ddagger}$, Matthew R. Short$^{1,3,\ast,\dagger,\S}$,
Lorenzo Vianello$^{1,\ast,\dagger}$,\and Daniel Ludvig$^{3}$, Levi Hargrove$^{1, 2, 3}$,  Kevin Lynch$^{2,4}$, Jose Pons$^{1,2,3,4,5}$\and
\small{$^1$ Shirley Ryan AbilityLab, Chicago, IL, USA}\and
\small{$^2$ Center for Robotics and Biosystems, Northwestern University, Evanston, IL, USA}\and
\small{$^3$ Department of Biomedical Engineering, Northwestern University, Evanston, IL, USA}\and
\small{$^4$ Department of Mechanical Engineering, Northwestern University, Evanston, IL, USA}\and
\small{$^5$ Department of Physical Medicine and Rehabilitation, Northwestern University, Chicago, IL, USA}\and
\small$^\ast$Corresponding author. Email:\and
\small baris.kucuktabak@gmail.com, matthewshort35@gmail.com, lvianello@sralab.org\and
\small{$^\dagger$ These authors contributed equally and are listed in alphabetical order.}\and
\small{$^\ddagger$ Current address: Honda Research Institute USA, San Jose, CA, USA}\and
\small{$^\S$ Current address: Department of Biomedical Engineering, University of Delaware, Newark, DE, USA}
}

\begin{document} 

\maketitle

\begin{abstract} \bfseries \boldmath
Following a stroke, individuals often experience mobility impairments due to weakness and loss of independent joint control in the lower limbs. As a result, gait recovery becomes a primary goal of physical rehabilitation, traditionally achieved through high-intensity therapist-led training.
However, conventional therapist-led approaches involving manual assistance or resistance can be physically demanding and limit interaction at multiple joints simultaneously.
Robotic exoskeletons have emerged as a promising solution, enabling multi-joint support, reducing therapist strain, and offering objective performance feedback. However, typical exoskeleton control strategies limit the physical therapist's involvement and adaptability to the patient’s needs, which may hinder clinical adoption and outcomes.
In this study, we introduce a gait rehabilitation paradigm based on physical Human-Robot-Human Interaction which we call Therapist-Exoskeleton-Patient Interaction (TEPI), in which a therapist and a post-stroke patient are each equipped with a lower-limb exoskeleton virtually connected at the hips and knees via spring-damper elements. This connection enables bidirectional physical interaction, allowing the therapist to guide the patient's movement while receiving real-time haptic feedback. We evaluated this approach with eight chronic post-stroke patients using a within-subject design, comparing TEPI training to conventional therapist-guided mobilization during treadmill walking. 
Results showed that TEPI led to greater joint range of motion, increased step length/height, and similar muscle activation compared to conventional therapy, as well as high self-reported motivation and enjoyment. These findings suggest that TEPI can integrate robotic precision with therapist intuition, offering a framework for enhancing gait rehabilitation outcomes in post-stroke populations.

\textcolor{black}{Summary: Therapist-Exoskeleton-Patient Interaction (TEPI) enables real-time shaping of post-stroke gait via bidirectional feedback.}
\end{abstract}

\section*{Introduction}

Following a stroke, individuals often experience mobility and balance impairments due to loss of independent joint control and strength deficits in the lower extremities \cite{arene2009understanding,sanchez2017lower}. During physical rehabilitation, gait recovery is a major focus for these individuals, given the importance of ambulation in activities of daily living at home and in the community \cite{rudberg2021stroke}. Current interventions primarily focus on high-intensity gait training administered by physical therapists \cite{belda2011rehabilitation}. In these sessions, therapists provide hands-on support and corrections to the limbs of the patient during overground or treadmill walking.

This approach requires substantial physical effort from the therapist, however, which can lead to work-related injuries over time \cite{mccrory2014work}. 
Additionally, because therapists must use their hands to guide and correct the patient's movements, they are limited in their ability to provide full-body cues and assessments, potentially reducing their ability to apply timely corrective forces and maintain engagement during gait therapy.
Moreover, one therapist can only physically interact with a patient at a few contact points simultaneously; as a result, the therapist's guidance is often isolated to a single aspect of the patient's gait, such as limb advancement. For this reason, the combined effort of multiple therapists is required for complex functional training involving the coordination of multiple joints, particularly in the early stages of rehabilitation \cite{stephan2021mobility}.

Emerging technologies, such as lower-limb exoskeletons, have been developed to address some of these challenges in gait training by enabling intensive multi-joint rehabilitation. These solutions can provide assistance or resistance to the patient at multiple contact points, reduce the number of therapists required during therapy, and decrease the therapist's effort by sustaining the patient's weight, while also providing objective measures of gait performance~\cite{belda2011rehabilitation, moeller2023use, gassert2018rehabilitation}. 
Among the various exoskeleton control strategies for gait restoration, the most common approaches include assist-as-needed and error augmentation techniques. The assist-as-needed method offers adjustable support levels tailored to the patient's ability, guiding the patient's limbs to follow predefined joint trajectories. This approach is especially beneficial for severely impaired patients in the early stages of rehabilitation, as it can encourage (re)learning of asymptomatic walking patterns~\cite{hobbs2020review, baud2021review, belda2011rehabilitation}. Conversely, error augmentation deliberately introduces errors or resistance during training, making it better suited for higher-functioning patients. By increasing awareness of gait deviations, this approach encourages adaptation and self-correction~\cite{marchal2019haptic, Marchal-Crespo2009}.

Defining optimal reference limb trajectories for exoskeleton controllers and appropriately adjusting assistance or resistance levels during therapy is challenging and remains an open question~\cite{belda2011rehabilitation, zhang2017human}. These control strategies often under-utilize the expertise of the physical therapist, limiting their influence on the behavior of the exoskeleton during training~\cite{hasson2023neurorehabilitation}.  
Commonly, reference trajectories are computed offline based on the walking patterns of able-bodied individuals~\cite{baud2021review}, which fail to adapt adequately to the patient’s performance during training and across various ambulatory activities. As a result, exoskeleton-based therapy reduces the therapist's ability to provide nuanced corrections to the patient’s movement, as they would in conventional gait therapy, potentially hindering the functional effects of such training as well as the adoption of this technology in clinical settings~\cite{celian2021day}. Furthermore, because therapists typically rely on external observation, either directly through visualization of limb kinematics or indirectly through performance metrics, their ability to engage in embodied, intuitive guidance is limited, diminishing real-time adaptability and interaction. This underscores the need for a balanced approach that combines the benefits of exoskeleton-based therapy with the expertise of the physical therapist to enhance rehabilitation outcomes for stroke survivors.

Using robots to mediate physical interaction between humans, known as physical Human-Robot-Human Interaction (pHRHI), provides unique opportunities to merge human motor-cognitive skills with robotic capabilities \cite{Kucuktabak2021}. In this paradigm, two individuals interface with separate robotic devices that are programmed to virtually render a physical interaction medium. Typically, this medium is modeled with springs and dampers, virtually linking either the joint configurations or the end-effectors of the robots.
As a result, each individual can feel and respond to their partner’s movements through the forces generated by the virtual connection.
Many studies have explored this paradigm during upper-limb \cite{waters2024theradyad, Ganesh2014, Takagi2017, Beckers2020, Takagi2018HapticInteraction, AvilaMireles2017SkillInteraction, Kager2019TheTask, Baur2019} and lower-limb \cite{Kim2021, Kim2023, Short2023, short2025effects, Kucuktabak2023, Vianello2024, koh2021exploiting} interaction tasks, reporting promising results in terms of enhanced motor performance and increased engagement during the interaction. 
\textcolor{black}{However, utilizing lower-limb exoskeletons to facilitate therapist-patient interaction during functional tasks, such as treadmill walking, has received limited attention.}

Virtually connecting a physical therapist to their patient using exoskeleton systems could be a promising alternative to existing gait training paradigms. This approach has the potential to harmonize the expertise and intuition of therapists with the precision and consistency of robots in real-time. By establishing a compatible interaction between the patient's and therapist's joints, such as virtually connecting the therapist's knee to the patient's knee, therapists can directly perceive joint-level deviations associated with gait impairments, enabling more informative assessments and more effective, phase-specific guidance during physical therapy.

\begin{figure*}[t!]
\centering
\includegraphics[width = 1.0\columnwidth]{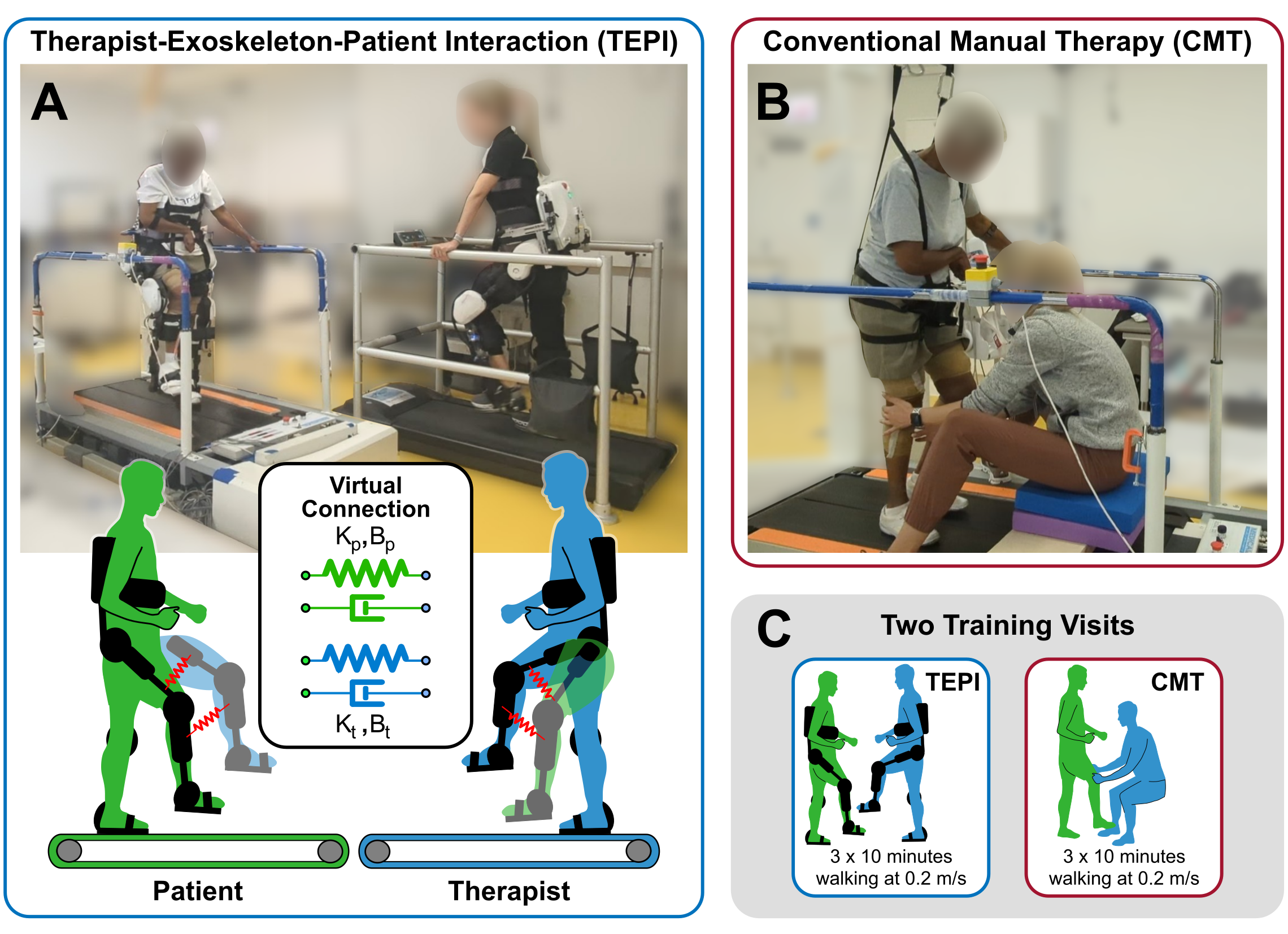}
\caption{Therapist-Exoskeleton-Patient Interaction (TEPI) for gait therapy: (A) a patient with chronic stroke and a physical therapist each wear lower-limb exoskeletons during a session of treadmill walking. The two exoskeletons are virtually connected via a spring-damper system that provides haptic feedback to both users. The patient and therapist face one another, facilitating visual, auditory and haptic communication between the two. The TEPI approach was compared to (B) conventional manual therapy (CMT) featuring manual, hands-on assistance from the physical therapist across two training sessions conducted on separate days. (C) Each session consisted of 30 minutes of treadmill training, divided into three blocks. The visit order was randomized and balanced across the eight participants (four TEPI-first, four CMT-first).}
\label{fig:dyad_setup}
\end{figure*}

Combining the capabilities of lower-limb exoskeletons with pHRHI, we introduce and evaluate a gait training approach: Therapist-Exoskeleton-Patient Interaction (TEPI)\textcolor{black}{, illustrated in Figure~\ref{fig:dyad_setup}}. In this study, we validate the proposed approach by using two lower-limb exoskeletons to mediate the physical interaction between a therapist and a post-stroke patient during treadmill walking. The controllers of the two exoskeletons render compliant spring-damper elements between the joints of the two users, allowing the therapist to guide the patient and receive haptic feedback of the patient's movement. 

We hypothesize that enabling a therapist to physically interact with a post-stroke patient through a virtual, joint-level connection between lower-limb exoskeletons will more effectively shape spatiotemporal gait characteristics than manual guidance alone, while maintaining patient engagement. Demonstrating robust, within-session modulation of spatiotemporal gait features, while preserving active participation and motivation, is a necessary precursor to evaluating learning, retention, transfer, and clinical carry-over for a new gait-training approach. Accordingly, this study focuses on within-session training performance.

The efficacy of the TEPI approach was evaluated in eight chronic stroke participants and compared to conventional manual therapy (CMT), where the therapist is seated and manually mobilizes the patient's joints. Each patient attended two sessions on separate days, one for each therapy type, consisting of 30 minutes of treadmill walking. Observations indicated that the TEPI resulted in a larger range of motion, including longer steps and larger step height, as well as increased muscle activation compared to CMT. At the same time, patients reported high levels of motivation and enjoyment during TEPI training, with similar levels of self-perceived exertion compared to CMT.

\section*{Results}

\subsection*{\textcolor{black}{Evaluation of TEPI and CMT during gait training}}
Eight post-stroke patients participated in two separate visits involving TEPI and CMT. The order of the two visits was randomized and equally balanced across participants, with four completing TEPI first and four completing CMT first.
In the TEPI visit, both the therapist and the patient wore lower-limb exoskeletons, with virtual springs and dampers rendered between their hip and knee joints. The stiffness parameters of the virtual connection, reported in Table S1, were determined on a subject-specific basis according to the physical therapist’s assessment of each patient’s needs; these parameters were adjusted as needed during the training sessions.
Our framework \textcolor{black}{allowed} us to independently change the spring-damper parameters for the patient's exoskeleton ($K_p$, $B_p$) and the therapist's exoskeleton ($K_t$, $B_t$), thus modifying the amount of torque feedback each partner perceives from one another. The range of virtual stiffnesses implemented was $K_p$ $\in$ [49, 64]~Nm/rad for the patient and $K_t$ $\in$ [25, 64]~Nm/rad for the physical therapist. Damping was selected proportional to the square root of the stiffness to maintain a constant damping ratio and avoid oscillatory behaviors.

Some desired outcomes of a therapy session for post-stroke patients include increasing range of motion, increasing step length and step height of the paretic limb, and maintaining high intensity (effort) during the session. We investigated how these metrics varied between TEPI and CMT sessions, where patients walked on a treadmill set to 0.2~m/s for three 10-minute blocks while receiving each type of training intervention. 

Joint kinematics were used to assess the ankle's workspace area, step length, and step height for both legs, as well as to measure the therapist's influence on the patient's trajectory. In the context of TEPI training, this influence was evaluated by examining the differences between the patient’s and the therapist’s joint movements. To measure effort, we recorded bipolar electromyography (EMG) from five patients, collected from four muscles on each leg: rectus femoris (RF), biceps femoris (BF), tibialis anterior (TA), and medial gastrocnemius (MG). EMG signals from these muscles were normalized with respect to free walking (without therapist intervention) and provided an indication of muscle activation relative to baseline walking patterns. Additionally, we used heart-rate sensors and a subjective questionnaire based on the \textcolor{black}{Borg} perceived rating of exertion scale \cite{borg}. To assess the patients' perceived difficulty and motivation associated with the training types, we provided survey questions adapted from the Intrinsic Motivation Inventory (IMI) \cite{mcauley1989psychometric} at the end of each training session.

\subsection*{Therapist-patient kinematic similarity during TEPI training}
\begin{figure}
\centering
\includegraphics[width = 0.9\columnwidth]{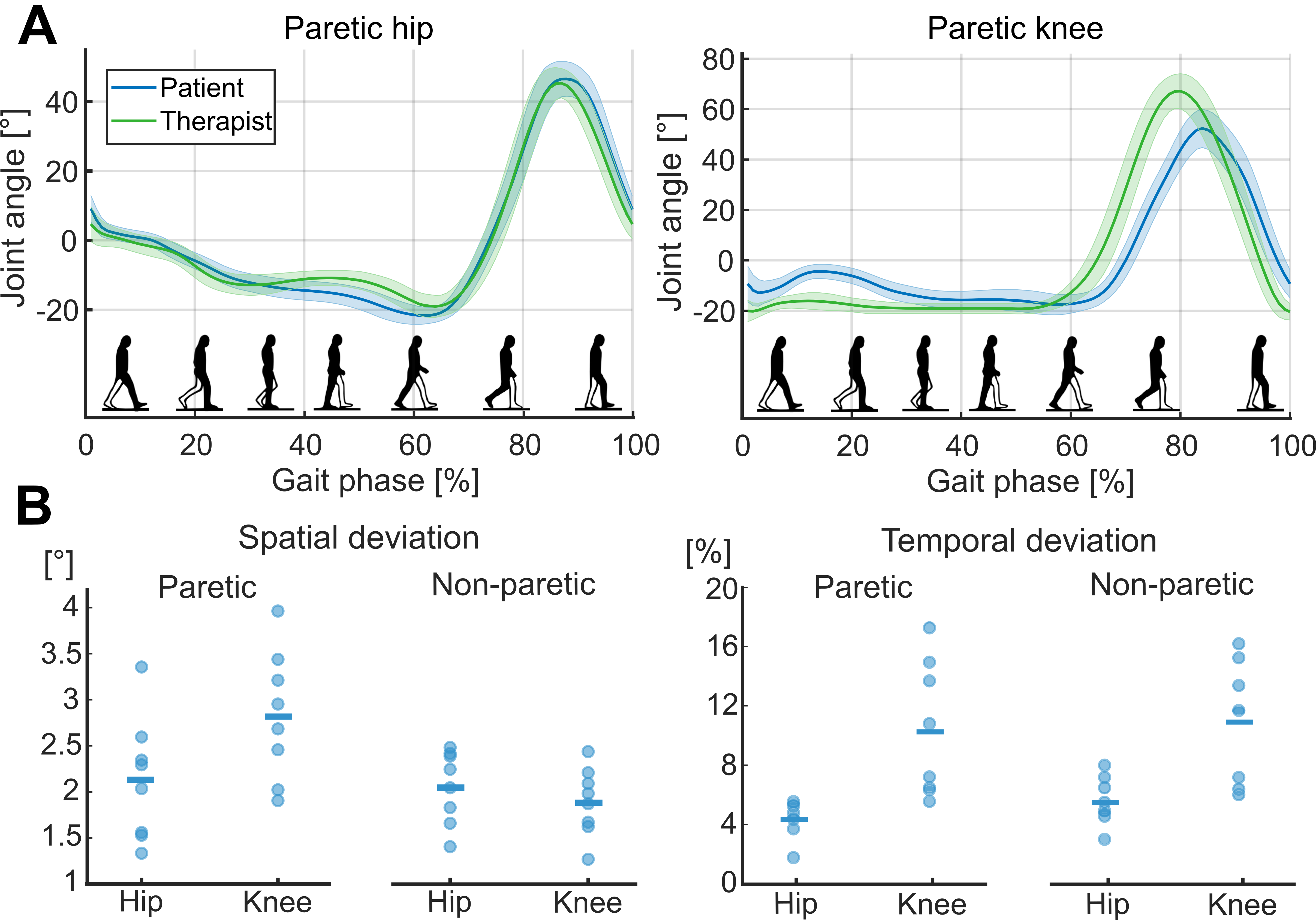}
\caption{\textbf{Spatiotemporal similarity between patient and therapist kinematics during TEPI training.} (\textbf{A}) Representative hip and knee joint angles for the patient and therapist during a TEPI training block, time-normalized to the gait cycle. Solid lines indicate mean across strides within a training block while shaded regions indicate mean $\pm$ one standard deviation. (\textbf{B}) Spatial and temporal deviation between the joint kinematics of the therapist and patient during TEPI session. Spatial deviation (left) was quantified as the root-mean-square-error between the joint configurations of the therapist and patient, after time-warping kinematics for each stride. Temporal deviation (right) was quantified as time-warped lag (percentage of gait cycle) between the therapist and patient joint kinematic time-series for each stride.}
\label{fig:joint_angle_similarity}
\end{figure}

The effectiveness of the TEPI approach was validated by computing the spatiotemporal similarity between the therapist’s and patient’s exoskeleton joint angles during training, confirming that the approach allowed the therapist to influence the patient's kinematics.
An example of these joint angles during a training block, normalized in time with respect to the gait cycle, is presented in Figure~\ref{fig:joint_angle_similarity} for a representative patient. During the TEPI session, the patient's trajectory closely \textcolor{black}{followed} the therapist's, with similar ranges of motion and temporal characteristics. For all patients and all joints, the mean spatial difference between therapist and patient was less than 4$^{\circ}$ (averaged across training blocks). On average, a larger spatial difference on the paretic side was observed compared to the non-paretic side. Regarding temporal differences, larger differences in alignment were observed at the knee joints (approximately 10\% of gait cycle) compared to the hip (5\% of gait cycle) for all patients.

\subsection*{Effects of TEPI on joint kinetics}
\begin{figure}
\centering
\includegraphics[width = 1.0\columnwidth]{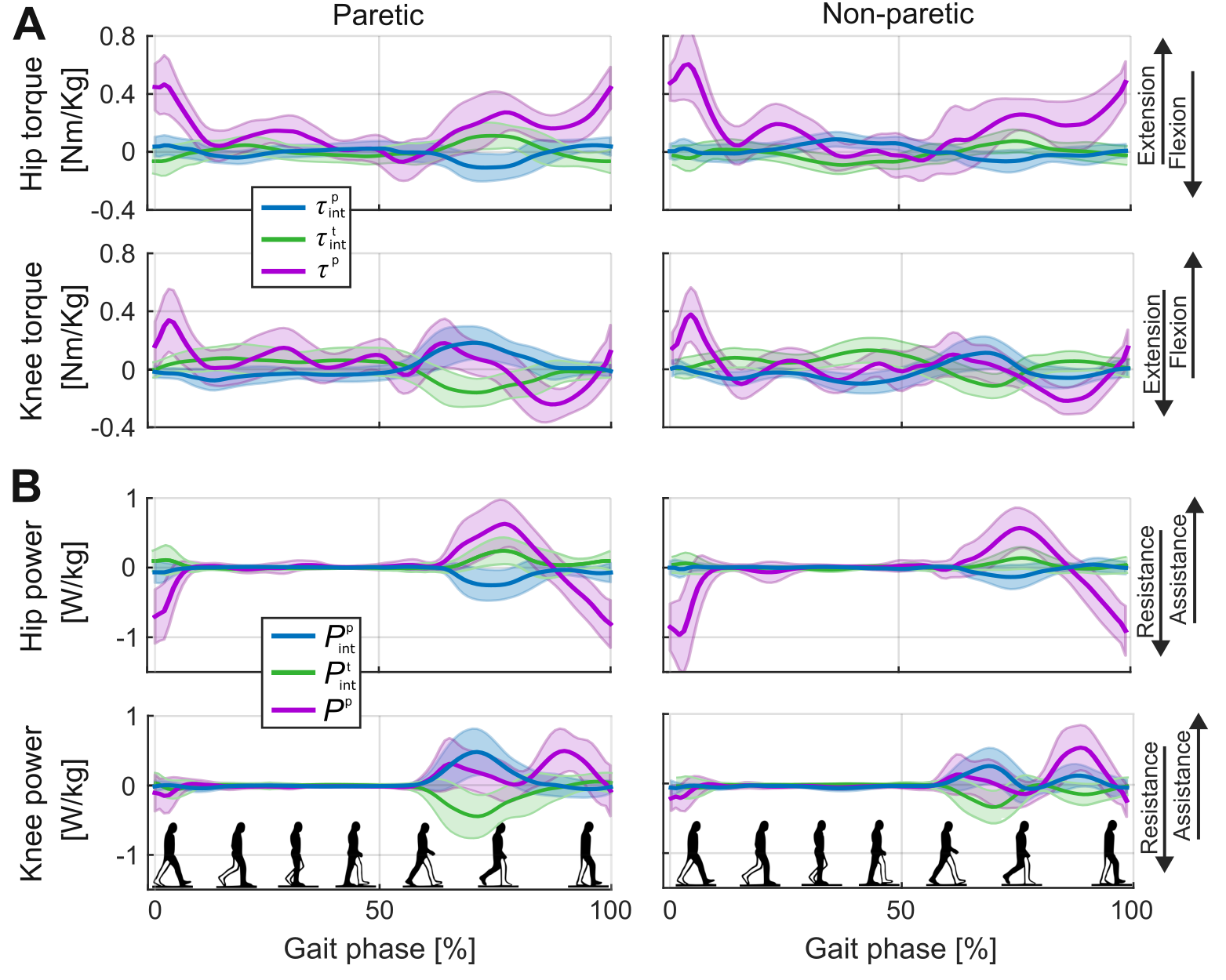}
\caption{\textbf{Joint kinetics during TEPI training.} Several relevant kinetic measurements are displayed across the normalized gait cycle, plot relative to the patient's paretic (left) and non-paretic (right) legs. Solid lines indicate the mean across strides, patients, and training blocks, while shaded regions represent the mean $\pm$ one standard deviation. Measures include (\textbf{A}) hip and knee torque as well as (\textbf{B}) hip and knee power. Patient interaction torque ($\tau_{int}^{p}$) and power ($P_{int}^{p}$) measured by the exoskeleton are shown in blue, therapist interaction torque ($\tau_{int}^{t}$) and power ($P_{int}^{t}$) are shown in green, and patient biological torques ($\tau^{p}$) and power ($P^{p}$) are shown in purple.}
\label{fig:joint-kinetics}
\end{figure}

{

To assess how TEPI influences joint kinetics, we analyzed joint torques and power during TEPI walking using exoskeleton interaction mechanics and inverse dynamics estimates \textcolor{black}{(Figure~\ref{fig:joint-kinetics})}. Specifically, we report exoskeleton interaction torques and corresponding interaction power at the hip and knee for both limbs, together with the patient’s net joint torques and joint power estimated from inverse dynamics.

In general, paretic and non-paretic joint kinetics were symmetric across the gait cycle. After time-warping each stride, joint torque magnitude and timing differences between limbs were small for the hip (error: 0.09 $\pm$ 0.03 Nm/kg, lag: 9.0 $\pm$ 5.3\% of the gait cycle) and knee (error: 0.07 $\pm$ 0.03 Nm/kg, lag: 6.7 $\pm$ 3.4\% of the gait cycle).

Peak biological joint torques were quantified by comparing limb-specific flexion and extension peaks. At the hip, peak flexion torque was higher for the paretic limb by 0.2 $\pm$ 0.2 Nm/kg. Other peak differences were smaller magnitude: peak hip extension torque was higher for the non-paretic limb by 0.01 $\pm$ 0.1 Nm/kg, peak knee flexion torque was higher for the non-paretic limb by 0.002 $\pm$ 0.2 Nm/kg, and peak knee extension torque was higher for the paretic limb by 0.04 $\pm$ 0.1 Nm/kg.

During swing, the paretic hip exhibited net resistive interaction (interaction power: $-$0.1 $\pm$ 0.07 W/kg), which was larger in magnitude than the non-paretic hip ($-$0.05 $\pm$ 0.06 W/kg). In contrast, during swing the paretic knee received net assistive interaction (interaction power: 0.2 $\pm$ 0.08 W/kg), again larger than the non-paretic knee (0.1 $\pm$ 0.08 W/kg). Stance-phase interaction was comparatively small for both joints and limbs (mean interaction power approximately 0.009 W/kg), indicating that the exoskeleton contribution was concentrated in swing rather than stance.

Importantly, even during phases of assistance, the patient remained an active contributor. During paretic knee swing, the biological torque accounted for 58\% of the total torque (biological and exoskeleton assistance combined); similarly, during non-paretic knee swing, biological torque accounted for 59\% of the total torque, indicating that swing initiation was not generated solely by the exoskeleton.
}

\subsection*{Effects of TEPI training on spatial gait measures}
\begin{figure}
\centering
\includegraphics[width = 1.0\columnwidth]{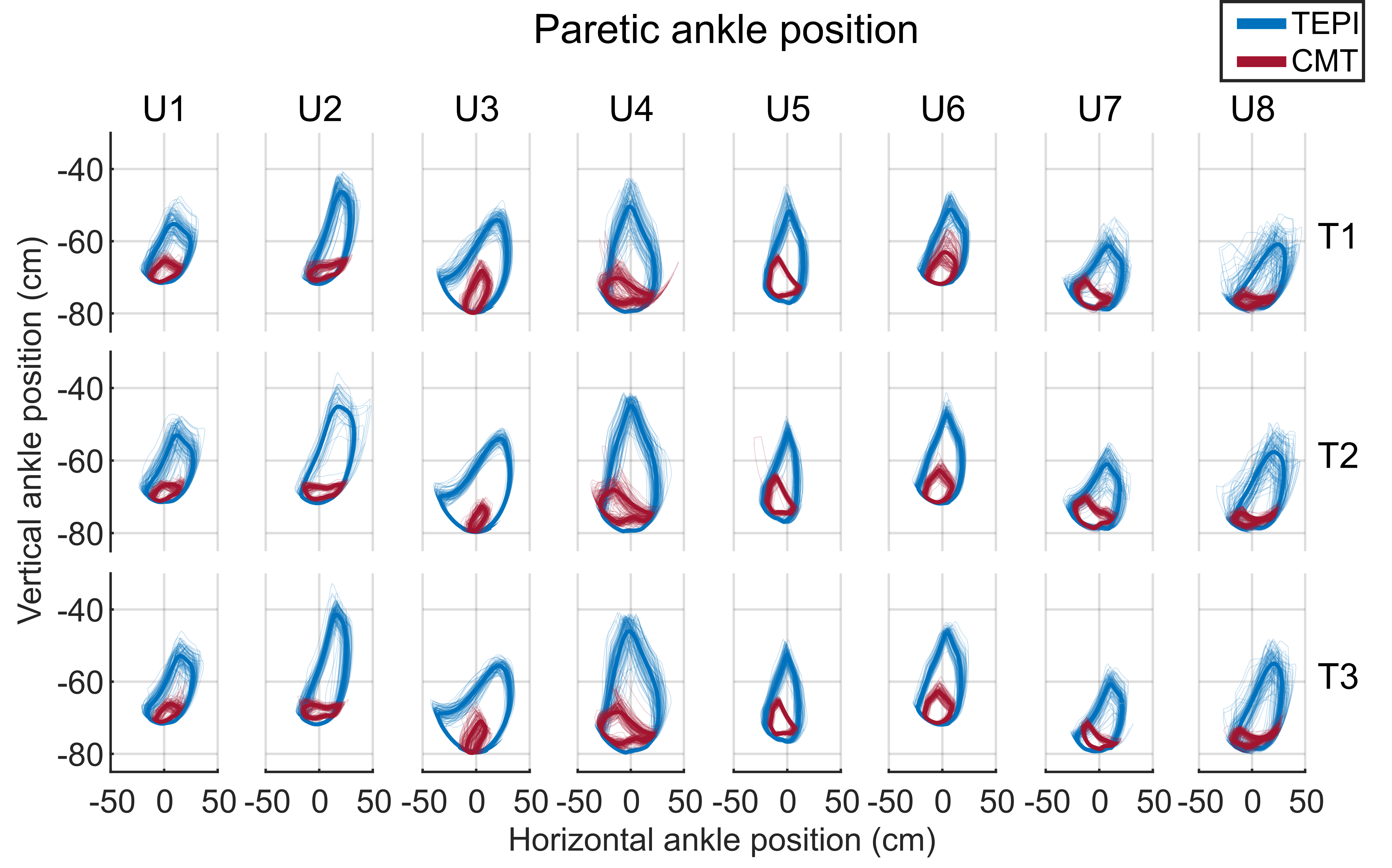}
\caption{\textbf{Paretic ankle trajectories during training.} For each patient (U1-U8), paretic limb ankle trajectories across training blocks (T1-T3) are displayed for the two training types: TEPI (blue) and CMT (red). Thin lines indicate individual strides while thick lines indicate the mean trajectory for a patient in a block of a training type.}
\label{fig:paretic-ankle}
\end{figure}

\begin{figure}
\centering
\includegraphics[width = 0.85\columnwidth]{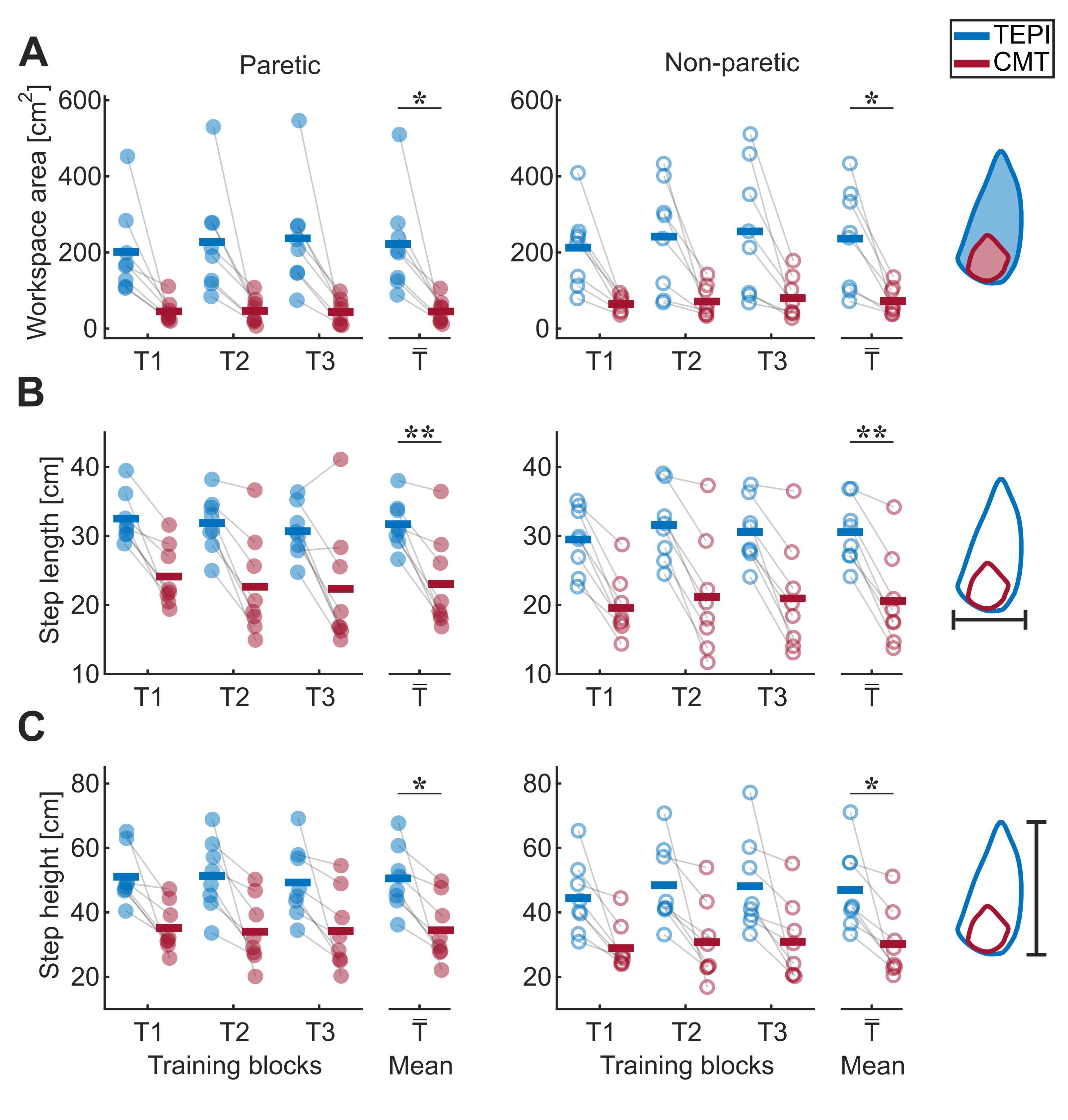}
\caption{\textbf{Spatial gait features during training.} Measures extracted from patients' sagittal plane ankle trajectories are shown for the paretic (left) and non-paretic (right) legs across training blocks (T1-T3), compared between the two training types: TEPI (blue) and CMT (red). (\textbf{A}) Workspace area indicates the total area of the 2-dimensional ankle trajectory, averaged across steps within each training block. (\textbf{B}) Step length indicates the horizontal distance between the landing limb and the stance limb. (\textbf{C}) Step height indicates the maximum vertical distance of the swing limb with respect to the stance limb. The mean of the training blocks ($\bar{\text{T}}$) was used to statistically compare the two training types using paired t-tests. * and **  indicate $p<$ 0.05 and 0.01, respectively, after Holm-Bonferroni correction (N = 8).}
\label{fig:ankle-spat}
\end{figure}

To compare the effects of the TEPI training with respect to CMT, we analyzed several measures related to the spatial characteristics of patients' gait during training. To do so, we used joint angles obtained from IMU sensors placed on the thigh and shank to calculate the Cartesian coordinates of the non-paretic and paretic ankles; from these ankle trajectories, we extracted multiple parameters to compare between training conditions: workspace area, step length, and step height. Ankle trajectories of the paretic limb are shown in Figure~\ref{fig:paretic-ankle}, displayed for each patient and block of training for the TEPI and CMT conditions. Group data for workspace area, step length, and step height are presented for the eight patients across training blocks in Figure~\ref{fig:ankle-spat}.

Workspace area \cite{sukal2006use}, related to the range of motion of patients during each training intervention, was considerably larger during TEPI compared to CMT, for both the paretic (TEPI: 221.9 $\pm$ 132.1 $\text{cm}^2$, CMT: 44.9 $\pm$ 32.2 $\text{cm}^2$; $t_{7}$ = 3.6, $p$ = 0.03) and non-paretic (TEPI: 236.5 $\pm$ 133.5 $\text{cm}^2$, CMT: 71.8 $\pm$ 36.9 $\text{cm}^2$; $t_{7}$ = 4.0, $p$ = 0.02) legs. Across the three training blocks, this increased workspace area was relatively constant, indicating that patients were able to successfully follow the physical therapist's joint configurations for the duration of the TEPI training. We further analyzed the step length (horizontal ankle position) and step height (vertical ankle position) and obtained similar trends comparing the two training conditions. On average, step length was larger in the TEPI condition compared to CMT, for both the paretic (TEPI: 31.7 $\pm$ 3.5 $\text{cm}$, CMT: 23.0 $\pm$ 6.8 $\text{cm}$; $t_{7}$ = 5.0, $p<$ 0.01) and non-paretic (TEPI: 30.5 $\pm$ 4.6 $\text{cm}$, CMT: 20.6 $\pm$ 6.8 $\text{cm}$; $t_{7}$ = 7.8, $p<$ 0.01) legs. Step height was also larger in the TEPI session, for both the paretic (TEPI: 50.5 $\pm$ 10.0 $\text{cm}$, CMT: 34.4 $\pm$ 10.0 $\text{cm}$; $t_{7}$ = 3.6, $p$ = 0.02) and non-paretic (TEPI: 47.0 $\pm$ 12.7 $\text{cm}$, CMT: 30.2 $\pm$ 10.6 $\text{cm}$; $t_{7}$ = 3.3, $p$ = 0.01) legs. For both training types, there was a trend of higher step length and step height for the paretic leg, compared to the non-paretic leg.

\subsection*{Effects of TEPI training on effort}
\begin{figure}
\centering
\includegraphics[width = 0.8\columnwidth]{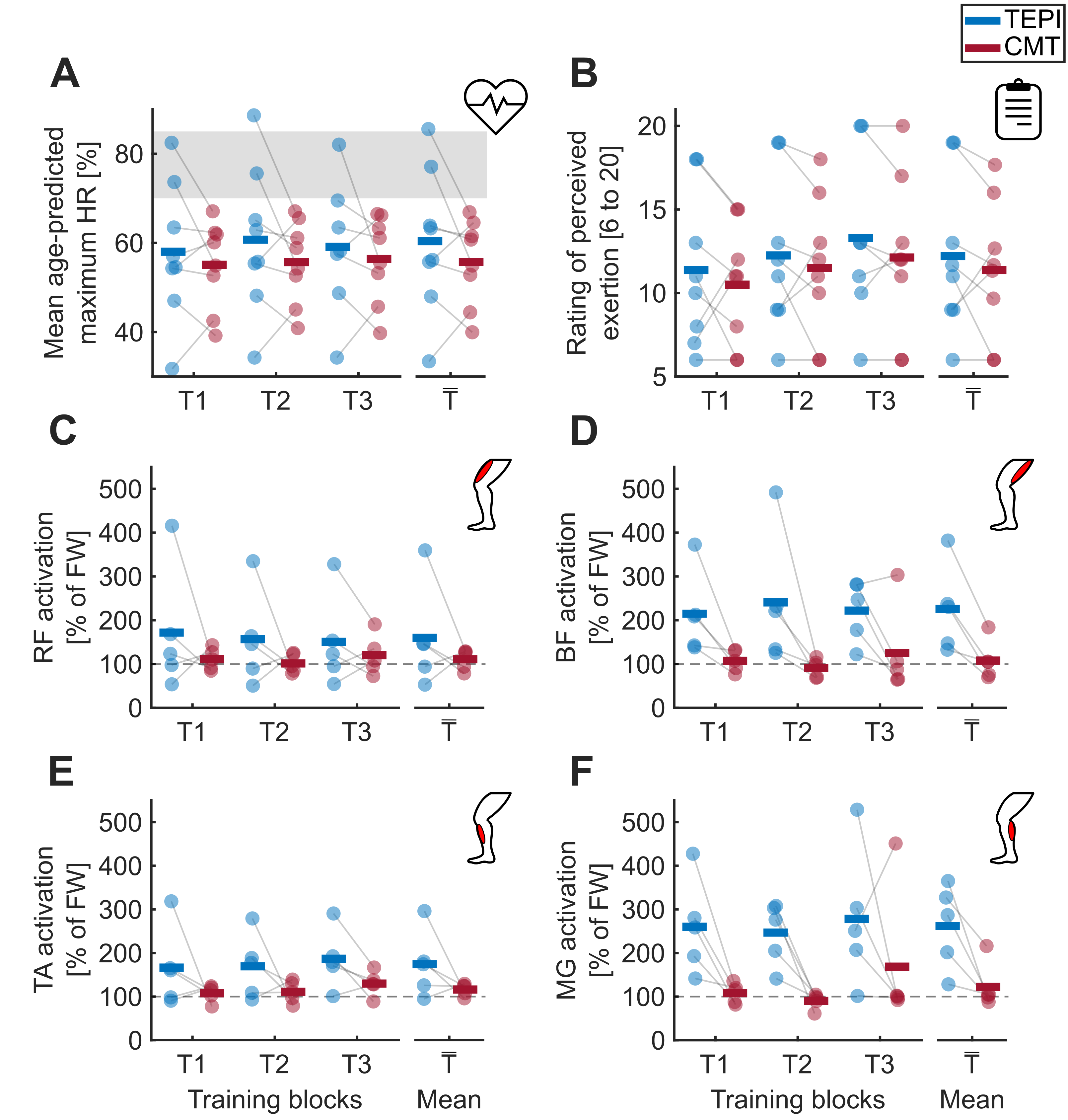}
\caption{Measures of patient effort across training blocks (T1-T3), compared between the two training types: TEPI (blue) and CMT (red). (Top Left) Mean percentage of age-predicted maximum heart rate (HR). The grey shaded region represents the target HR range during training. (Top Right) Rating of perceived exertion (RPE) showing subjective levels of effort during training. Higher values indicate greater perceived effort. (Bottom) Mean muscle activation in the paretic limb, with respect to free walking (no training intervention). Each point represents a patient's mean activation. 100\% indicates the mean muscle activation during free walking. The mean of the training blocks ($\bar{\text{T}}$) was used to statistically compare the two training types using paired t-tests (N = 8 for HR and RPE, N = 5 for muscle activation).
}
\label{fig:combined_effort}
\end{figure}

To evaluate each patient's effort during training, we analyzed mean heart rate, self-perceived measures of exertion, and muscle activation. The mean heart rate (HR) for each block within a session, as a percentage of the age-predicted maximum \cite{tanaka2001age}, was calculated using the formula:

\begin{equation}
\text{HR} = 208 - 0.7 \times \text{age},
\label{eq:heart_rate}
\end{equation}

and is visualized in the top panels of Figure~\ref{fig:combined_effort}.

During TEPI training, we observed that patient heart rate was higher than during CMT (TEPI: 60.4 $\pm$ 16.3\%, CMT: 55.7 $\pm$ 9.6\%; $t_{7}$ = 1.1, $p$ = 0.29), however this difference was not statistically significant. Furthermore, in both conditions, heart rate during training was less than the target heart rate zone during therapy (70-85\% of age-predicted maximum heart rate) for a majority of the patients. Heart rate findings were supported by the patients' perceived exertion on the 6-20  \textcolor{black}{Borg} scale. Patients reported higher scores on the perceived effort scale for the TEPI training (12.2 $\pm$ 4.7) compared to CMT (11.4 $\pm$ 4.2), though this difference was not statistically significant ($t_{7}$ = 0.84, $p$ = 0.43). Notably, both heart rate and perceived rate of exertion tended to increase from the first to the last training block, indicated by our repeated measures ANOVA which revealed a significant effect of block number on heart rate ($F_{2,12}$ = 7.8, $p$ $<$ 0.01) and perceived exertion ($F_{2,12}$ = 14.1, $p$ $<$ 0.001); however, the lack of a significant interaction between block number and training type suggests this increase in heart rate and reported exertion was similar for both the TEPI and CMT conditions.

In terms of muscle activation (Figure~\ref{fig:combined_effort}), we observed trends of increases in the average muscle activity, normalized with respect to free walking, during TEPI training compared to CMT. Muscle activation did not change significantly across training blocks for either training type. Though there was considerable variance across patients, we observed slightly higher muscle activation for the RF (TEPI: 159.7 $\pm$ 118.3\%, CMT: 110.9 $\pm$ 23.2\%; $t_{4}$ = 0.93, $p$ = 0.4), BF (TEPI: 225.8 $\pm$ 99.2\%, CMT: 107.8 $\pm$ 45.5\%; $t_{4}$ = 4.0, $p$ = 0.07), TA (TEPI: 174.2 $\pm$ 76.7\%, CMT: 116.2 $\pm$ 13.9\%; $t_{4}$ = 1.5, $p$ = 0.4) and MG (TEPI: 261.7 $\pm$ 96.2\%, CMT: 122.2 $\pm$ 52.9\%; $t_{4}$ = 3.3, $p$ = 0.09). In general, muscle activation during TEPI training was higher than free walking ($>$ 100\%) while muscle activation during CMT training was closer to free-walking magnitudes. However, it is important to note that these comparisons were only made for a subset of our full cohort (U4-U8), and comparisons did not reach significance after corrections for multiple comparisons.

\subsection*{Effects of training on self-reported motivation and task demand}

\begin{figure}
\centering
\includegraphics[width = 1.0\columnwidth]{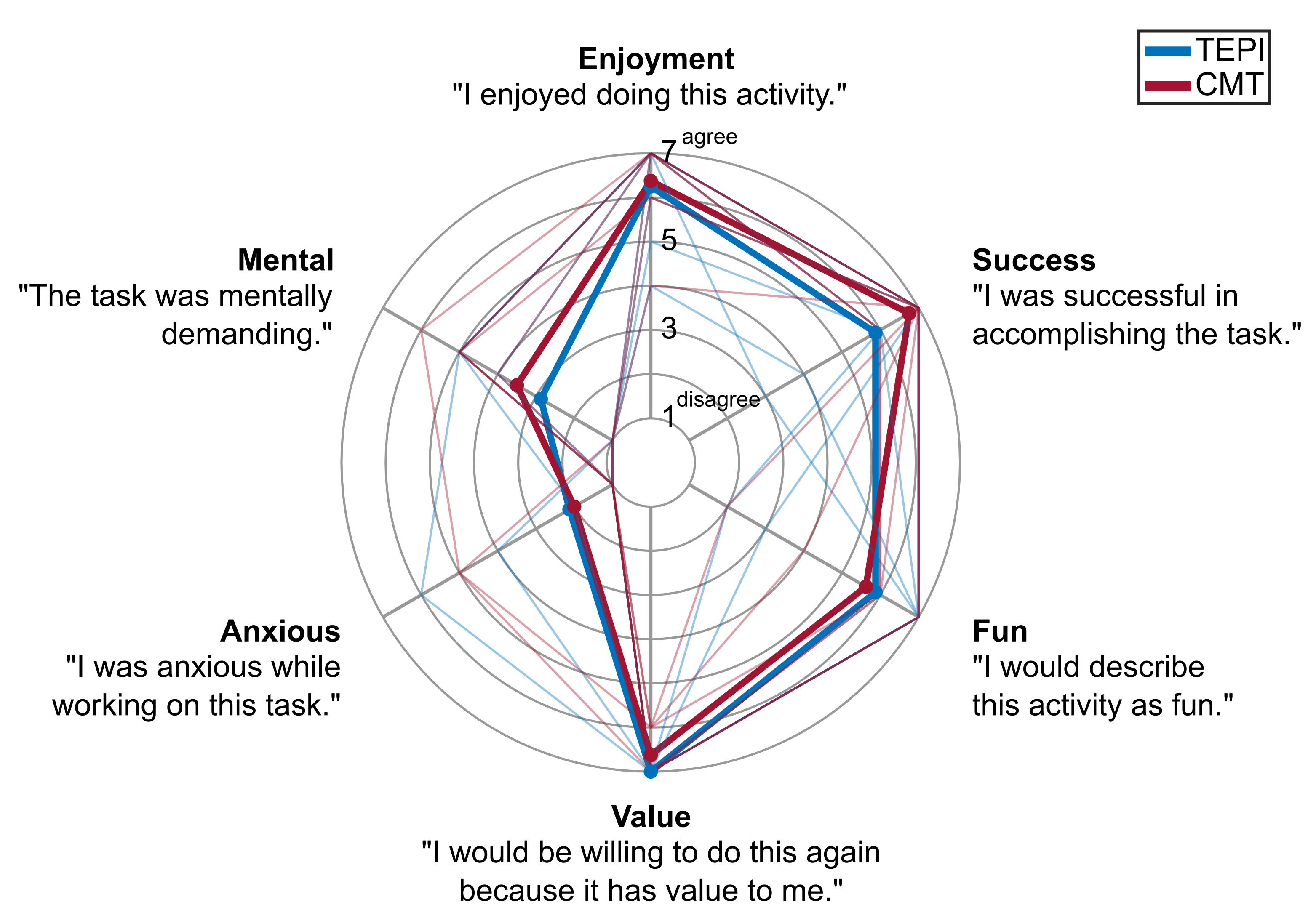}
\caption{\textbf{Patient survey responses post-training.} Spider plot shows responses to survey statements adapted from the Intrinsic Motivation Inventory (IMI) for each training type: TEPI (blue), CMT (red). Higher scores indicate agreement with the feelings/descriptions associated with each question. Wilcoxon signed-rank tests were used to compare patient responses across training types (N = 8).}
\label{fig:IMI}
\end{figure}

In general, patients reported similar results for IMI survey questions related to motivation, enjoyment, and perceived difficulty of the TEPI and CMT training (Figure~\ref{fig:IMI}). Using Wilcoxon signed-rank tests, we found no difference between training types for any of the questions answered. Patients indicated that they found the training valuable (TEPI: 7.0 $\pm$ 0.0, CMT: 6.6 $\pm$ 0.5; $p$ = 1.0) and had fun (TEPI: 5.9 $\pm$ 2.1, CMT: 5.6 $\pm$ 1.8; $p$ = 1.0) upon completion of both training types, with scores close to the highest agreeable response. Mental demand responses were moderate and comparable between training types (TEPI: 2.9~$\pm$~2.0, CMT: 3.5~$\pm$~2.1; $p$ = 1.0).
Other categories also showed no consistent trend across patients; feelings of anxiety (TEPI: 2.1 $\pm$ 1.9, CMT: 2.0 $\pm$ 1.9; $p$ = 1.0) during the training remained relatively low and were comparable between the training types.

\section*{Discussion}

\subsection*{\textcolor{black}{TEPI enables multi-contact therapist guidance of patient kinematics}}

During TEPI, the patient's trajectory closely followed the therapist's, exhibiting similar ranges of motion and temporal characteristics. A key advantage of TEPI training is its ability to provide forces at multiple contact points, enabling a synergistic correction of both hip and knee movements during walking. Additionally, this approach provides bilateral assistance to both the stance and swing legs, simultaneously addressing key features like knee buckling, swing initiation, and weight acceptance. Achieving comparable multi-contact, bilateral assistance with manual therapy would likely require multiple therapists, with the additional challenge of coordinating assistance timing among these clinicians.

In terms of kinematics during TEPI training (Figure S1), distinct behaviors were observed between the paretic hip and the knee joints of the patient. \textcolor{black}{Although} the patient's hip joint showed a slight temporal lead compared to the therapist's during the swing phase, the opposite relationship was observed for the knee joint. This suggests that the therapist may preferentially focus on the assistance of knee flexion to reduce compensatory strategies typically used by the patient, such as excessive hip circumduction \cite{chen2005gait,kerrigan1999spastic}. This interpretation is further supported by the kinematics of the non-paretic leg, where no \textcolor{black}{discernible} temporal lead was observed in either the hip or the knee joints, indicating a more natural and symmetric coordination in the absence of more \textcolor{black}{pronounced} impairments.

In addition, the patient demonstrated a smaller range of motion at the paretic knee compared to the therapist. This discrepancy is likely due to the increased joint stiffness of the patient's knee \cite{souissi2018co}, which may have prevented them from achieving the full range of motion imposed by the therapist, even when physical interaction was present. Such limitations are consistent with the general trend in post-stroke patients, where distal joints tend to be more affected than proximal joints \cite{lim2023functional}.


Temporal and spatial differences in joint kinematics resulted in variations in the magnitude of the interaction torques, as shown in Figure~\ref{fig:joint-kinetics} and Supplementary Figure S1. Notably, greater interaction torque was observed at the non-paretic hip around 40\% of the gait cycle, indicating increased therapist support to the stance leg during the swing phase of the paretic limb. Moreover, during the swing phase (60–100\%), increased interaction torques were observed at both the paretic hip and knee joints. 
These interaction torques translated into perceived assistance or resistance for the therapist and the patient, varying by joint and gait phase, as observed in the interaction power plots in Figure~\ref{fig:joint-kinetics}.
Overall, these findings suggest that the therapist delivered targeted, phase-specific corrections to both legs, implementing an interaction strategy that was relatively consistent across the post-stroke patients in this study cohort.

Two main factors affect the forces delivered to the patient: the virtual connection properties ($K_t, B_t, K_p, B_p$) and the joint-level deviation between therapist and patient kinematics. Shown in Table S1, we observed infrequent changes in the virtual connection properties across patients and training blocks. Therefore, it is likely that the therapist was able to comfortably shape the patient's gait at a selected set of connection properties by dynamically adjusting her own movements during the training. \textcolor{black}{Since} the virtual connection is implemented in joint space and the hardware is individually fitted, our framework accommodates substantial therapist–patient anthropometric differences (160–180 cm and 56–99 kg in our cohort).

It is important to note that the spatial and temporal similarities between patient and therapist kinematics do not imply that the patient was \textcolor{black}{substantially} limited or restricted in their ability to deviate from the therapist's guidance. To illustrate this, we present instantaneous time-series differences between patient and therapist joint angles without time-warping (Figure S3), capturing both temporal and spatial kinematic deviations. These instantaneous differences ranged from $-$15$^{\circ}$ to $+$25$^{\circ}$ across patients and joints, demonstrating the movement variability that is provided and inherent to our compliant TEPI framework.

\subsection*{\textcolor{black}{TEPI increases range of motion compared to conventional therapy}}

Compared to CMT, TEPI led to increased range of motion, step length, and step height in both the paretic and non-paretic legs. These findings suggest that the physical therapist was more effective in modifying patients' movements to resemble asymptomatic walking patterns during TEPI training compared to CMT. Interestingly, in both training conditions, step length and step height tended to be larger for the paretic leg than for the non-paretic leg. This suggests that the therapist deliberately exaggerated the movement of the paretic leg during training. This approach differs from conventional exoskeleton control strategies, which typically either assist the motion as needed to complete the movement, such as assist-as-needed, or amplify the patient's error, such as error augmentation.

\textcolor{black}{Although} different strategies might be adopted in longer training sessions or with different patients, this observation highlights the need to reconsider exoskeleton control approaches for gait rehabilitation. The optimal control strategy for promoting neuroplasticity in neurorehabilitation remains unclear. The proposed approach could help uncover the motor mechanisms therapists target when physically interacting with patients.
The holistic dataset collected during dyadic exoskeleton training, including kinematic, and dynamic information from both therapist and patient, is provided in the Supplementary Materials. This dataset can be leveraged by the community to identify therapist-driven control strategies and develop fully collaborative robotic controllers. 

\subsection*{TEPI training promotes active patient involvement}

To assess patients' effort during each training session, we analyzed mean heart rate, self-perceived effort, and muscle activation. Results indicated that heart rates were slightly higher during TEPI compared to CMT, though not significantly different; subjective perceived effort showed a similar trend. Notably, heart rate and ratings of perceived exertion tended to increase from the first to the last training block for both training types, potentially due to fatigue onset throughout the training sessions. Overall, heart rate and perceived exertion should be interpreted cautiously because they can reflect factors beyond mechanical work, including novelty of wearing an exoskeleton and harness, attentional demands, and being observed by the research staff. The laboratory and treadmill environment were the same for TEPI and CMT, but we acknowledge that the patient would likely view TEPI as more novel than CMT.

Muscle activation was generally higher during TEPI training compared to CMT, though not significantly different. Furthermore, EMG activation levels did not change across the TEPI training blocks, indicating sustained effort throughout the 30-minute session. \textcolor{black}{Although} it is possible that this increased muscle activation can be partially attributed to the constraints imposed by the exoskeleton (movement in the sagittal plane), these results suggest that the patients were not passive during the exoskeleton training. To contextualize the contribution of wearing the device itself, we additionally report EMG data adapted from our previous work \cite{Kucuktabak2024} with healthy participants (N = 3) in Supplementary Figure S4. Although this sample size is not sufficient for inferential statistics, the descriptive trends show that wearing the exoskeleton in transparent mode resulted in approximately 40 to 80\% higher mean EMG activation relative to free walking. In contrast, in our post-stroke cohort, mean EMG during TEPI (normalized to free walking without the exoskeleton) was approximately 60 to 160\% higher across muscles. Therefore, the elevated EMG activation observed during TEPI likely reflects meaningful patient effort during movement execution, not solely the burden of wearing the exoskeleton. In other words, the improved range of motion, step length, and step height reflected the combined effect of therapist-delivered robotic assistance and the patient’s active engagement in following that guidance.

To further evaluate whether TEPI supports active patient participation rather than passive movement, we examined joint kinetics during TEPI walking in addition to joint kinematics and spatial gait measures. The decomposition of biological and interaction torques indicates that assistance and resistance can be applied by the therapist in a phase- and joint-specific manner. For example, the resistive interaction at the paretic hip during late swing is consistent with the therapist resisting to reduce hip leading, whereas the assistive interaction at the paretic knee during swing initiation supports knee flexion for limb advancement. Importantly, even during assistive phases at the paretic knee, the patient still generated positive biological knee power, and the biological contribution remained a substantial portion of the total joint torque ($\sim$58\%), indicating that swing initiation was not produced solely by the therapist. Moreover, across the gait cycle, paretic and non-paretic biological joint torques showed only minor timing and magnitude differences, consistent with TEPI promoting coordinated, relatively symmetric limb loading during training. Together, these kinetic patterns support the interpretation that TEPI promotes active patient involvement while enabling the therapist to shape the patient's gait in a phase-specific manner.

\subsection*{Effects of training on self-reported motivation and task demand}

In general, patients reported similar results for IMI survey questions related to motivation and perceived difficulty of the TEPI and CMT sessions. Patients indicated that they enjoyed and had fun upon completion of both training types, with a slight preference towards the TEPI training in the categories of \textbf{Fun} and \textbf{Value} for most patients. In terms of cognitive load, a majority of patients reported that the TEPI training was less mentally demanding than the CMT. Other categories showed no consistent trend across patients; feelings of anxiety remained relatively low during the training and were comparable between the training types. Importantly, these survey results suggest that chronic stroke patients remain motivated throughout training and do not appear deterred by the constraints imposed by exoskeleton-based assistance, viewing it as comparable to conventional manual support.

\subsection*{Practical considerations and steps toward clinical adoption}

TEPI requires two exoskeletons and an actively engaged therapist, creating practical challenges to routine clinical use. These include streamlined donning and doffing, simpler therapist-facing interfaces, such as treadmill speed control and real-time virtual connection stiffness adjustment, improved system reliability and fault handling, as well as deployment-ready safety features. The cost of multiple exoskeletons is a practical constraint for many clinics until hardware prices decrease. Additionally, although the therapist in this study completed only 40 minutes of familiarization with two healthy participants, structured training and competency assessment may also \textcolor{black}{influence} the indirect costs of clinical adoption.

Optimizing the quality of a TEPI session is another key practical step toward clinical adoption before moving into multi-session motor learning studies. Building on the presented feasibility and benefits of TEPI, we identify several directions to enhance performance and enable patient-specific personalization of the virtual connection. First, a systematic evaluation of the interaction parameters ($K_t$, $B_t$, $K_p$, $B_p$) could characterize how these gains shape outcomes. Second, therapist-facing interfaces that permit intra-session stiffness adjustments, not only between blocks, can better align assistance with dynamic therapeutic goals. Third, semi- or fully autonomous real-time tuning/optimization with safety protections can help to avoid unsafe settings. In particular, sample-efficient, preference-based active learning~\cite{Lee2023} could offer a principled route to patient- and therapist-in-the-loop personalization, leveraging inputs from both users.

\subsection*{Implications for motor learning and clinical outcomes}

Although the present study was not designed to measure patients' adaptation following TEPI (changes in gait after training), previous studies in healthy individuals suggest that a single session of pHRHI-based training with a partner can improve individual adaptation during visuomotor tracking tasks. For example, Ganesh et al. \cite{Ganesh2014} showed that, during a 2-DoF reaching task with a visuomotor rotation, pHRHI training leads to faster individual learning rates with respect to solo training, highlighting the benefits of haptic communication during collaboration. Ivanova et al. \cite{ivanova2022interaction} found that training 1-DoF wrist movements with pHRHI, compared to training with static trajectory guidance, improved retention and transfer of tracking performance (accuracy and movement smoothness) immediately after training as well as one day and one week later. Moreover, in one of our recent studies, we compared single session adaptation effects associated with physical human interaction and trajectory guidance during a visuomotor ankle tracking task in a chronic stroke cohort \cite{short2025haptic}. In this study, we demonstrated potential short-term benefits associated with practicing 1-DoF ankle movements while virtually connected to a physical therapist, which preserved movement variability during training while leading to enhanced muscle activation immediately after training. These results indicate that compliant, partner-responsive interactions can facilitate learning during simple visuomotor tracking tasks in healthy individuals and post-stroke patients. It remains an open question, however, how patients post-stroke may benefit from TEPI gait training in terms of short-term individual learning or long-term functional recovery.

A key distinction between previous pHRHI-based learning studies and our TEPI framework is the type of interaction provided. Previous studies have focused solely on haptic interaction between pairs of individuals; these pairs performed tracking tasks simultaneously without explicit knowledge of the virtual interaction medium, controlling for psychosocial effects which may influence learning when interacting with another human. In our TEPI framework, we allowed the therapist and patient to interact not only through the haptic medium, but also through visual and auditory channels to facilitate an interactive training that was consistent with conventional exercises in neurorehabilitation. 
\textcolor{black}{Although prior work has examined haptic interaction and social interaction separately in motor learning and rehabilitation contexts, the combined effects of visual, auditory, and haptic interaction between peers during motor tasks remain less well understood.}
However, research on social interaction between partners, such as rehabilitation games with visual and auditory communication, can provide additional insights into the positive effects of incorporating such feedback modalities. Previous work has shown that post-stroke patients report high motivation and engagement during multi-player rehabilitation games and generally prefer interacting with a partner compared to single-player training \cite{baur2018robot, mace2017balancing, gorvsivc2018multisession}, attributed to social participation and increased agency while cooperating in group settings. High levels of motivation and engagement are important in the context of rehabilitative exercises, as explicit cognitive strategies require appropriate attentional resources and play a critical role in acquiring new skills and forming motor memories \cite{krakauer2019motor}. Therefore, we believe our TEPI framework has the potential to produce positive learning outcomes, due to a combination of the advantages of pHRHI training and the elevated motivation of social multi-player interaction.

In this work, TEPI enabled the therapist to modulate spatiotemporal features of the patient's gait, enlarging joint movements and spatial metrics during a single session of training while maintaining engagement and manageable effort. Demonstrating robust, within-session modulation is a necessary precursor to evaluating short- and long-term learning. Our results establish a foundation for longitudinal studies that test such changes in learning, and for design refinements that enhance reliability, safety, and therapist usability. 

\subsection*{Limitations}

\textcolor{black}{Although} the comparison of our TEPI framework and CMT is relevant in understanding the therapist's ability to modify their patient's kinematics during training, this comparison has limitations. For higher functioning individuals with chronic stroke, physical therapists would likely implement progressive, resistive strategies, as opposed to manual assistance, to more appropriately challenge patients during training \cite{ouellette2004high}. As a first step in the validation of our framework in a neurologically impaired population, we focused on higher-functioning individuals with chronic stroke and gave the physical therapist freedom to decide which specific aspects of gait to correct for each patient across visits. In addition, the proposed framework was tested during conventional, body weight-supported treadmill training. However, to better understand its \textcolor{black}{influence} and advantages, future research should explore more complex ambulatory activities.
For instance, assessing this framework during ambulatory activities of daily living, such as overground walking, stairs, and ramps, could provide valuable insight into how this technology may be adapted for clinical use. Moreover, since standardized clinical assessments of lower-limb function were not included in this study, future longitudinal studies should incorporate clinical measures such as the Lower Extremity Fugl-Meyer Assessment and overground gait biomechanics to evaluate training-related changes and long-term recovery.

In this study, we used an overground exoskeleton to enable extensions to overground walking and sit-to-stand tasks. This type of exoskeleton can produce relatively high torque output but has lower speed capabilities; however, the presented method could also be implemented with treadmill-based exoskeletons. Implementing the interaction force controller on such systems would be simpler, as they do not involve ground contact or support of the user’s full body weight~\cite{Kucuktabak2024}. Accordingly, training in this study was performed on a treadmill at a fixed conservative speed of 0.2 m/s, which was lower than our participants’ self-selected overground speeds, to minimize speed-related confounds in the analysis of joint movements and support safe, sustained interaction in the coupled therapist and patient configuration. Future studies should evaluate TEPI at faster speeds, including therapist-controlled and algorithmically adaptive speed progression with real-time safety monitoring, and test whether within-session effects on step length, step height, and joint movements generalize across a wider range of speeds consistent with each patient's functional capacity.

\textcolor{black}{Though} human gait primarily involves coordinated hip, knee, and ankle movements in the frontal and sagittal planes, the exoskeleton used in this study only provides sagittal plane actuation at the hip and knee joints. Expanding support to the frontal plane offers an opportunity to incorporate balance training alongside kinematics improvements. In addition, incorporating ankle assistance could enhance training of pre-swing push-off behaviors, a critical component of post-stroke gait rehabilitation \cite{hsiao2015mechanisms}. Using the general framework of TEPI, the integration of these additional degrees of freedom can be explored in future studies.

\section*{Conclusion}
In this study, we introduce a gait therapy approach in which a therapist and a post-stroke patient each wear lower-limb exoskeletons and train together through virtual physical interaction. This approach aims to harmonize the expertise and intuition of physical therapists with the capabilities of robots, providing real-time assistance based on the physical therapist's lower-limb kinematics and their visual, auditory, and haptic assessment of the patient's behaviors. We evaluated this approach with eight chronic stroke patients and a physical therapist, comparing results to conventional methods of manual assistance. Our results demonstrated that Therapist-Exoskeleton-Patient Interaction (TEPI) led to significantly greater range of motion, including larger step length and step height, and similar muscle activation compared to conventional manual assistance, as well as high levels of self-reported motivation and enjoyment. These findings suggest that incorporating physical human-robot-human interaction into therapeutic practices has the potential to improve the effectiveness of post-stroke gait rehabilitation; however, conducting longitudinal studies to quantify long-term effects and addressing practical challenges related to setup, interfaces, hardware burden, and enhanced safety mechanisms will be key steps toward broader clinical adoption.

\section*{Materials and Methods}

\subsection*{Participants and experimental protocol}
Eight patients (60.1 $\pm$ 15.2 years) with chronic hemiparetic stroke participated in this study. \textcolor{black}{Inclusion criteria for participating post-stroke patients were: diagnosed with unilateral, supratentorial, ischemic or hemorrhagic stroke at least six months prior to study recruitment, and ability to walk greater than 10 meters independently on level ground.} \textcolor{black}{Exclusion criteria were: body weight exceeding 100~kg, pregnancy, botulinum toxin injections to the lower limbs within the prior three months, current wounds or pressure ulcers, history of substance abuse, reduced cognitive function or severe aphasia, co-existing neurological or peripheral nerve disorders, severe lower-limb arthritis, recent fractures or osteoporosis, Modified Ashworth Scale greater than or equal to 3, and any medical or psychiatric condition that could interfere with study procedures.} Patient characteristics are detailed in Table~\ref{tab:participants}. The patients gave their written consent in accordance with the Declaration of Helsinki. The study protocol (STU00212684) was approved by the Institutional Review Board of Northwestern University \textcolor{black}{and registered on \url{clinicaltrials.gov} (NCT04578665)}.
The same licensed physical therapist (175 cm, 61 kg) conducted both sessions for all patients. Prior to the study experimentation, the therapist was trained to use the TEPI platform in two familiarization sessions, each consisting of a 20-minute block of treadmill walking while virtually connected to a healthy participant.

\begin{table}
\centering
\begin{tabular}{c| c c c c c c c} 
\cline{1-8}
\rule{0pt}{2ex}  
 Patient & Sex & Age & Height  & Body weight & Paretic & Time since  & SSW Speed  \\
   &  & [years] & [cm]  & [kg] & Side  &   stroke [years] &  [m/s]  \\\cline{1-8} 
  U1 & F & 76 & 160 & 56 & Right & 21 & 0.67 \\ 
  U2 & M & 66 & 163 & 66 & Right & 14 & 0.58 \\ 
  U3 & M & 70 & 178 & 85 & Right & 21 & 0.77  \\ 
  U4 & M & 73 & 175 & 90 & Left & 6 & 0.91  \\ 
  U5 & F & 65 & 165 & 99 & Right & 6 & 0.69  \\ 
  U6 & M & 55 & 170 & 74 & Left & 15 & 1.07  \\
  U7 & F & 34 & 170 & 91 & Right & 4 & 1.10  \\
  U8 & M & 42 & 180 & 79 & Left & 3 & 0.69  \\ \cline{1-8}
\end{tabular}
\caption{\textbf{Patient characteristics.} Eight patients with chronic hemiparetic stroke participated in the two visits of this study. Self-selected walking (SSW) speed was collected during a 10-meter walk test on the first visit as a measure of each patient's gait capacity.}
\label{tab:participants}
\end{table}

Each patient attended two training visits (TEPI and CMT) separated by a minimum of one week; the order of these visits was randomized across patients. Before each session of training, the patient performed three minutes of treadmill walking at 0.2 m/s without wearing an exoskeleton and without receiving any assistance from the therapist to obtain their baseline walking pattern. During this pre-assessment, patient kinematics and muscle activations were recorded using EMG/IMU sensors (Trigno Avanti, Delsys). To minimize speed-related confounds in the analysis of joint movements and to prioritize safety, treadmill speed was fixed at 0.2 m/s for all participants during both training conditions.

Prior to the TEPI sessions, the exoskeleton’s link lengths were individually adjusted for each patient. Following this adjustment, participants completed a brief familiarization with and without the virtual connection while standing, involving movements of one leg at a time. Following the familiarization phase, three 10-minute training blocks were conducted, with approximately 3-minute breaks between each session. During the training blocks, the patient and therapist faced one another, and the therapist provided physical assistance through the virtual connection of the two exoskeletons, in addition to auditory cues. Between blocks of the TEPI training session, the therapist had the flexibility to modify the amount of haptic feedback from the patient (by changing $K_t$ in Eq. \ref{eq:joint_space}) to provide easier control when necessary. Table S1 displays the stiffness for each training block and each user. Due to a technical issue with the exoskeleton, one patient (U2) was only able to complete two of the three training blocks.

During the CMT sessions (Figure~\ref{fig:dyad_setup}), the therapist sat in front of the treadmill and provided manual assistance to the patient, consistent with conventional body weight-supported treadmill training \cite{tarihci2024effectiveness}. The therapist had the freedom to interact with one or both legs, depending on her assessment of the patient's needs. Auditory cues were again provided by the therapist to facilitate the training. To provide a visual aid similar to the TEPI approach, a mirror was placed in front of the patients during CMT training.

\subsection*{\textcolor{black}{Exoskeleton control and virtual connection}}

\begin{figure}
\centering
\includegraphics[width = 0.9\columnwidth]{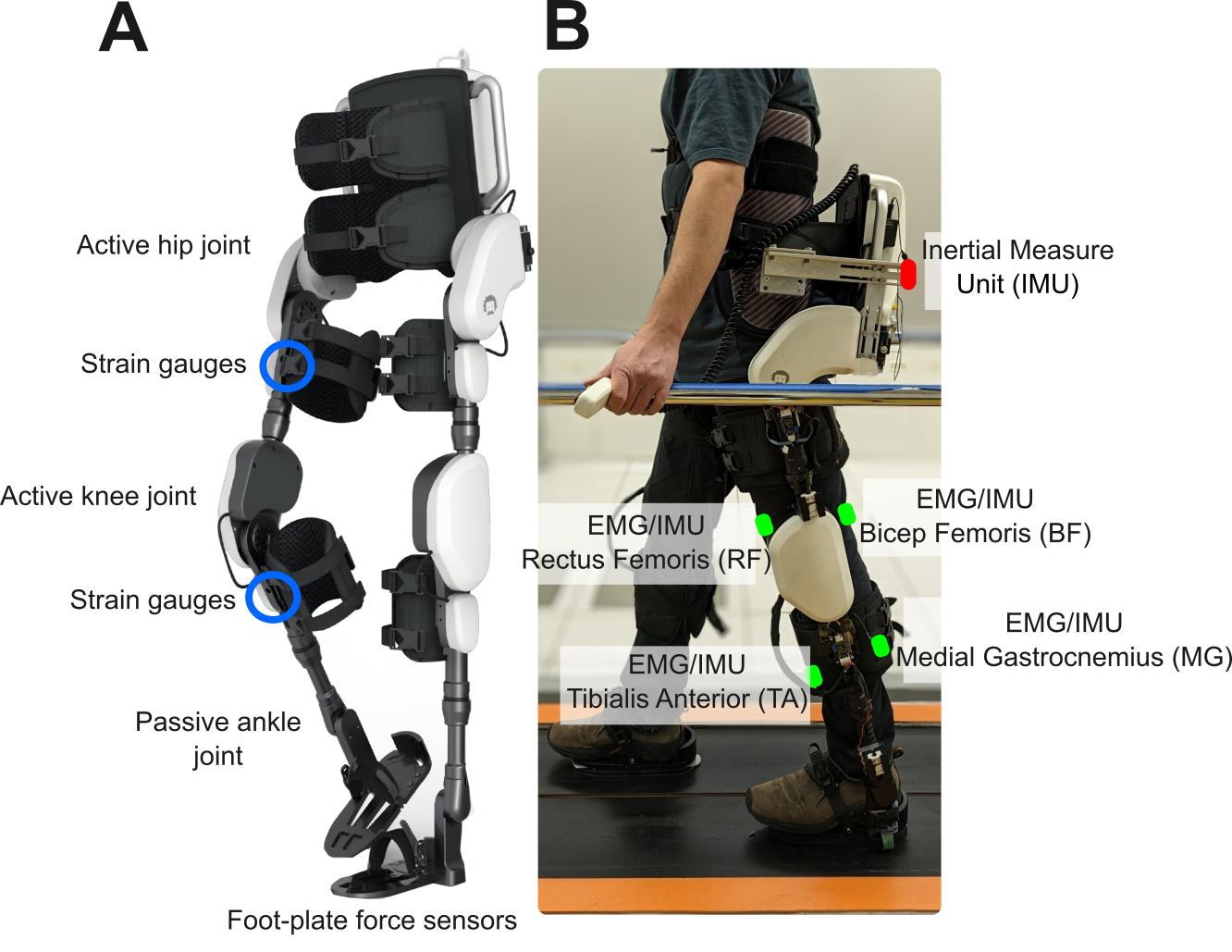}
\caption{\textbf{Exoskeleton user setup.} (\textbf{A}) A Fourier Intelligence ExoMotus-X2 lower-limb exoskeleton is shown. We highlight key components of the exoskeleton, including four active degrees-of-freedom (DoFs) at the hip and knee joints, two passive DoFs at the ankle joints, strain gauges on the thigh and shank to measure interaction forces, and force sensors on the feet to monitor ground contact. (\textbf{B}) A user wears the exoskeleton while the EMG and IMU sensor placements implemented in both experimental conditions (TEPI, CMT) are visualized.}
\label{fig:x2}
\end{figure}

Two lower-limb exoskeletons (ExoMotus-X2, Fourier Intelligence), illustrated in Figure~\ref{fig:x2}, were modified and implemented in our TEPI framework for gait therapy. The X2 exoskeleton, designed for overground walking, features four active degrees-of-freedom (DoFs) at the hip and knee joints, while passively supporting the ankle joints with a torsional spring.

Communication with the motors and onboard sensors was facilitated through the CAN bus using CANOpen protocol. To minimize the communication lag, an external PC was tethered to both exoskeletons. The control systems were developed using a ROS and C++ based open-source software framework, CANOpen Robot Controller (CORC)~\cite{FongCanOpenDevelopment}, and operated at a frequency of 333 Hz on the external PC.

To render the virtual interaction medium between the two exoskeletons, we implemented torsional spring and damper elements between the hip and knee joints of the therapist and the patient. These elements \textcolor{black}{determined} the amount of haptic feedback perceived by the therapist and patient, and \textcolor{black}{could} be adjusted independently for each user. As the therapist and patient \textcolor{black}{faced} one another during training, their opposite legs \textcolor{black}{were} virtually coupled. In other words, the left and right legs of the therapist \textcolor{black}{were} coupled with the right and left legs of the patient, respectively; this was implemented in alignment with mirror therapy approaches commonly used during gait training \cite{li2018effects}. Based on the properties of the virtual interaction elements and the instantaneous joint states of both exoskeletons ($t$: therapist's exoskeleton, $p$: patient's exoskeleton), desired interaction torques for each exoskeleton were calculated for each joint as
\begin{align}
     \hat{\tau}_t^* &= K_t(\tilde{\theta}_p - \theta_t) + B_t(\dot{\tilde{\theta}}_p - \dot{\theta}_t), \\
     \hat{\tau}_p^* &= K_p(\tilde{\theta}_t - \theta_p) + B_p(\dot{\tilde{\theta}}_t - \dot{\theta}_p),
     \label{eq:joint_space}
\end{align}
where $\hat{\tau}_i^*:i \in {t, p}$ are the desired interaction torques for the exoskeleton of the therapist and the patient, $(K_i, B_i) : i \in {t, p}$ are the stiffness and damping components, $(\theta_i, \dot{\theta}_i)$ and  $(\tilde{\theta}_i, \dot{\tilde{\theta}}_i):i \in {t, p}$ are the joint angles and velocities of the two opposite legs. Desired interaction torques for each joint of each user were commanded to their corresponding exoskeleton~\cite{Kucuktabak2023}.

Both exoskeletons \textcolor{black}{ran} independent instances of a whole-exoskeleton closed-loop interaction torque controller~\cite{Kucuktabak2024} to track desired interaction torques determined by the virtual interaction elements. \textcolor{black}{This} controller \textcolor{black}{compensated} for whole-body dynamics, including the full weight of the exoskeleton, and \textcolor{black}{calculated} the interaction torque error during the entire gait cycle. \textcolor{black}{Interaction torque error was used as an input to a virtual admittance model with desired joint acceleration as the output}. Joint accelerations \textcolor{black}{were} then fed into an optimization scheme to follow these commands under physical and safety constraints. Compensating for the undesired weight and inertial forces of the exoskeletons \textcolor{black}{allowed} the therapist and patient to accurately perceive one another's forces through the rendered virtual interaction medium.

\subsection*{Data recording and processing}

Exoskeleton kinematic data were measured at 333 Hz through encoder readings of the hip and knee joints. Additionally, wireless EMG/IMU sensors (Trigno Avanti, Delsys) were used to measure muscle activation and kinematics during both visits. Eight sensors were placed on the bellies of the following muscles in both legs: rectus femoris (RF), biceps femoris (BF), tibialis anterior (TA), and medial gastrocnemius (MG). For CMT, an additional IMU was placed on the trunk; for TEPI, an IMU sensor on the backpack of the exoskeleton was used.

EMG data were sampled at 1000~Hz. Initially, the raw EMG signals were bandpass filtered between 20 and 500~Hz using a sixth-order Butterworth filter. Subsequently, a notch filter was applied between 59 and 61~Hz to minimize noise from power-line interference. After full-wave rectification, the signals were low-pass filtered at 5~Hz. The filtered quaternion output of the Trigno sensors, sampled at 148~Hz, was used to measure joint angles in the sagittal plane. All data related to kinematics and muscle activation were windowed and time-normalized between consecutive heel-strikes, detected by ground reaction force sensors instrumented under a split-belt treadmill.

Heart rate (HR) was measured using optical heart rate sensors (Verity Sense, Polar). Additionally, after each training block, each patient reported their self-perceived rating of exertion (RPE) using the \textcolor{black}{Borg} scale~\cite{borg}.




\textcolor{black}{We adapted six questions from the IMI \cite{mcauley1989psychometric} to assess the self-reported difficulty and motivation associated with each training type. We asked patients to state their agreement or disagreement with six statements, scored from 1 to 7, where 1 indicated total disagreement and 7 indicated total agreement. The statements were: ``I enjoyed doing this activity'' (\textbf{Enjoyment}), ``I was successful in accomplishing the task'' (\textbf{Success}), ``I would describe this activity as fun'' (\textbf{Fun}), ``I would be willing to do this again because it has value to me'' (\textbf{Value}), ``I was anxious while working on this task'' (\textbf{Anxiety}), and ``The task was mentally demanding'' (\textbf{Mental}).}

\subsection*{Inverse dynamics}
{
We performed inverse dynamics analyses using OpenSim to estimate patient joint torques and joint power during TEPI walking. For each subject, a musculoskeletal model was generated by scaling a generic model according to the participant’s height and body mass. Joint kinematics were low-pass filtered at 6~Hz prior to inverse dynamics computations. The inverse dynamics output \textcolor{black}{represented} the net joint torque required to produce the measured motion given the external forces, which in our setup reflects the combined contributions of the patient’s biological joint torques and the exoskeleton interaction torques,
\begin{equation}
\tau_{ID} = \tau_{\text{human}} + \tau_{\text{interaction}}.
\end{equation}

To isolate the patient’s biological contribution, we subtracted the measured exoskeleton interaction torques from the inverse dynamics estimates. Joint power was computed as the product of joint torque and joint angular velocity, representing the instantaneous rate of mechanical energy generation or absorption at each joint. All torque and power measures were normalized by body weight. Because interaction forces between therapist and patient were not measured during the CMT condition, these inverse dynamics decompositions were performed for TEPI trials only.
}

\subsection*{\textcolor{black}{Statistical analysis}}
A two-way, repeated measures ANOVA was used to assess the effect of training block number (T1, T2, T3) and training type (TEPI, CMT) on each of the reported outcomes measured across blocks (spatial gait features, muscle activation, HR, RPE). ANOVA results are reported in Tables S2-S4. For comparisons related to spatial gait and muscle activation, the training block number, and its interaction with training type, were non-significant, leading us to collapse across conditions and focus on comparisons between the mean of the training blocks ($\bar{\text{T}}$). As these were our primary outcome measures, we used a similar approach when analyzing our secondary outcome measures of HR and RPE for consistency in reporting. Shapiro-Wilk tests confirmed that the mean differences between conditions were normally distributed for most outcome measures ($p>$ 0.05); paired t-tests were used to compare these normally distributed outcome measures between the two training visits: TEPI and CMT. Wilcoxon signed-rank tests were used to compare the survey results between TEPI and CMT, as these results were not normally distributed. Holm-Bonferroni corrections to reported $p$-values were performed to control for family-wise error rate within each of our statistical outcomes (paretic/non-paretic spatial gait measures, paretic muscle activation, HR, RPE, survey questions). Significance was set to 0.05 for all statistical tests. Results are provided as mean $\pm$ standard error unless otherwise specified.

\section*{Supplementary Materials and Methods}
\noindent\textbf{The PDF file includes:}
\begin{itemize}
    \item Dataset description
    \item Figs. S1 to S4
    \item Tables S1 to S4
\end{itemize}

\noindent\textbf{Other supplementary material for this manuscript includes:}
\begin{itemize}
    \item Movie S1
    \item Data file S1
\end{itemize}

\clearpage 

%

%
%


\section*{Acknowledgments}
We would like to thank Clément Lhoste, Hamidollah Hassanlouei, Laura Bandini, Francesco Di Tommaso, Abhishek Sankar, and Jialu Yu for their help with experimental preparations and data collection, Grace Hoo for recruiting patients, Tim Haswell for his technical support on the hardware improvements of the ExoMotus-X2 exoskeleton, Francesco Lanotte for the setup pictures, Sangjoon Kim and Yue Wen for their assistance with the force controller development, and Justin Fong and Vincent Crocher for initiating and leading the open-source CanOpenRobotController.
\paragraph*{Funding:}
This work was supported by the National Science Foundation~/~National Robotics Initiative (Grant No: 2024488).
\paragraph*{Author contributions:}
E.B.K. contributed to conceptualization, infrastructure, and experimental design, conducted the initial experiments, developed software for data analysis, verified the data, and wrote and revised the manuscript. M.R.S. and L.V. contributed to experimental design, conducted all experiments, performed data analysis and verification, and wrote and revised the manuscript. D.L. supervised the experimental design and statistical analyses and revised the manuscript. L.H., K.L., and J.L.P. provided funding, supervised the experimental design and manuscript preparation, and revised the manuscript.
\paragraph*{Competing interests:}
There are no competing interests to declare.
\paragraph*{Data and materials availability:}
All data needed to evaluate the conclusions in the paper are present in the paper or the Supplementary Materials. The data for this study have been deposited in the  Dryad database:\\ (\url{https://datadryad.org/share/kkuygTn_rEXdvFofV5540gZoehvy2s-4RX-x2y8KmP8}). \textcolor{black}{No new materials were generated in this study.}




\newpage


\renewcommand{\thefigure}{S\arabic{figure}}
\renewcommand{\thetable}{S\arabic{table}}
\renewcommand{\theequation}{S\arabic{equation}}
\renewcommand{\thepage}{S\arabic{page}}
\setcounter{figure}{0}
\setcounter{table}{0}
\setcounter{equation}{0}
\setcounter{page}{1} 


\begin{center}
\section*{Supplementary Materials for\\ \scititle}

Emek Barış Küçüktabak$^{\ast,\dagger}$, Matthew R. Short$^{\ast,\dagger}$,
Lorenzo Vianello$^{\ast,\dagger}$,\\ Daniel Ludvig, Levi Hargrove, Kevin Lynch, Jose Pons\\
\small$^\ast$Corresponding author. Email:\\
\small baris.kucuktabak@gmail.com, matthewshort35@gmail.com, lvianello@sralab.org\\
\small{$^\dagger$ These authors contributed equally and are listed in alphabetical order.}
\end{center}

\subsubsection*{This PDF file includes:}
Dataset description\\
Figs. S1 to S4\\
Tables S1 to S4\\

\subsubsection*{Other Supplementary Materials for this manuscript:}
Movie S1\\
Data file S1\\

\newpage

\subsection*{Dataset description}

The associated dataset is available as supplemental material and can be accessed through the provided link (\url{https://datadryad.org/share/kkuygTn_rEXdvFofV5540gZoehvy2s-4RX-x2y8KmP8}). 
Included with the dataset are sample MATLAB scripts and a README.md file to guide readers on how to use the data (\url{https://github.com/ponsLab/Th-Exo-Patient-pI}).

\begin{figure}
    \centering
    \includegraphics[width=0.9\linewidth]{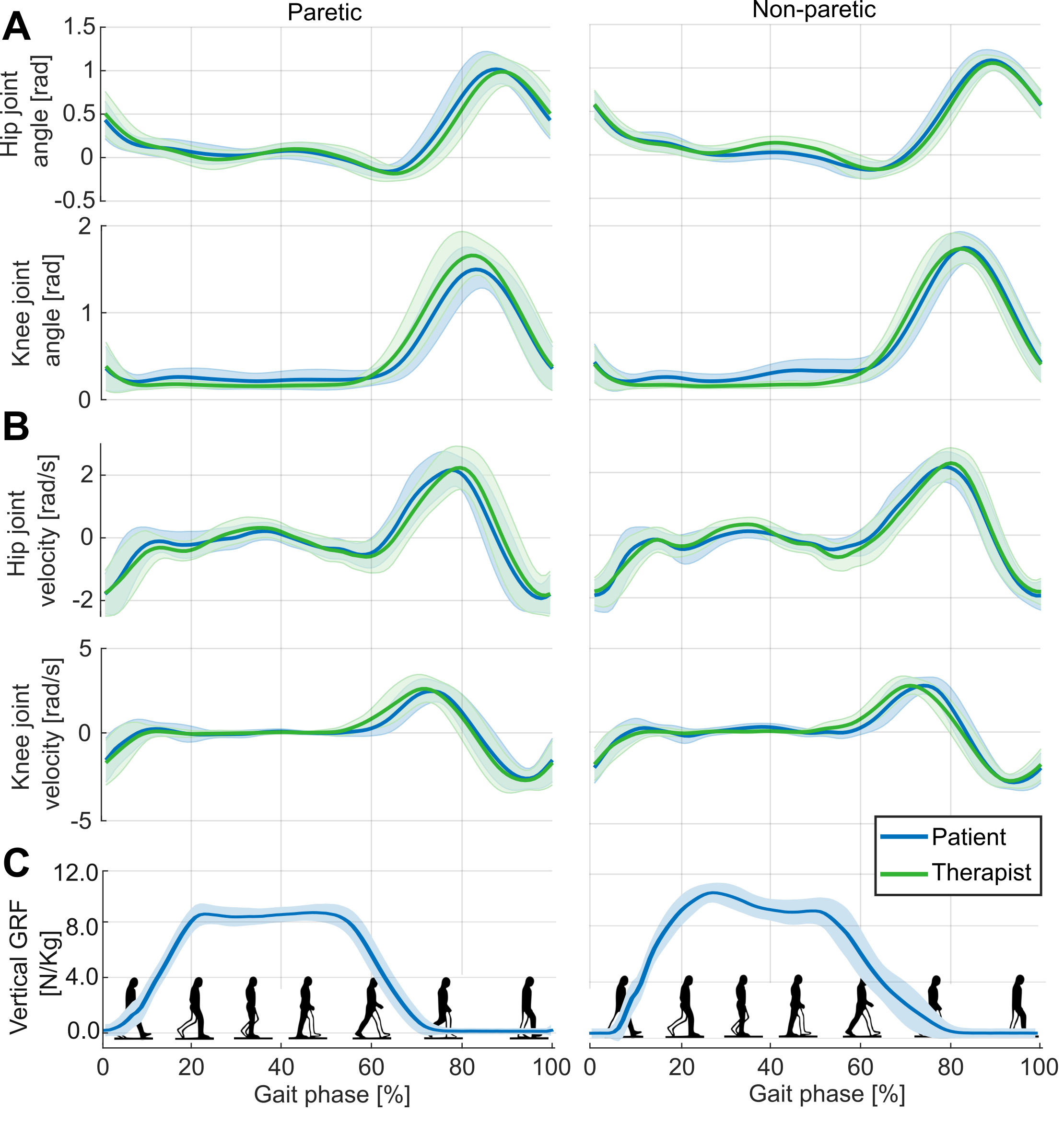}
    \caption{\textbf{Patient-therapist kinematics and ground reaction forces during TEPI training.} Measures are shown for the paretic (left) and non-paretic (right) legs with solid lines indicate the mean across strides, patients, and training blocks, while shaded regions represent the mean $\pm$ one standard deviation. From top to bottom, the plots show hip joint position, hip joint velocity, knee joint position, knee joint velocity, and vertical ground reaction force (GRF). Patient data are shown in blue, while therapist data are shown in green.}
    \label{fig:sup-kin}
\end{figure}

\begin{figure}
\centering
\includegraphics[width = 1.0\columnwidth]{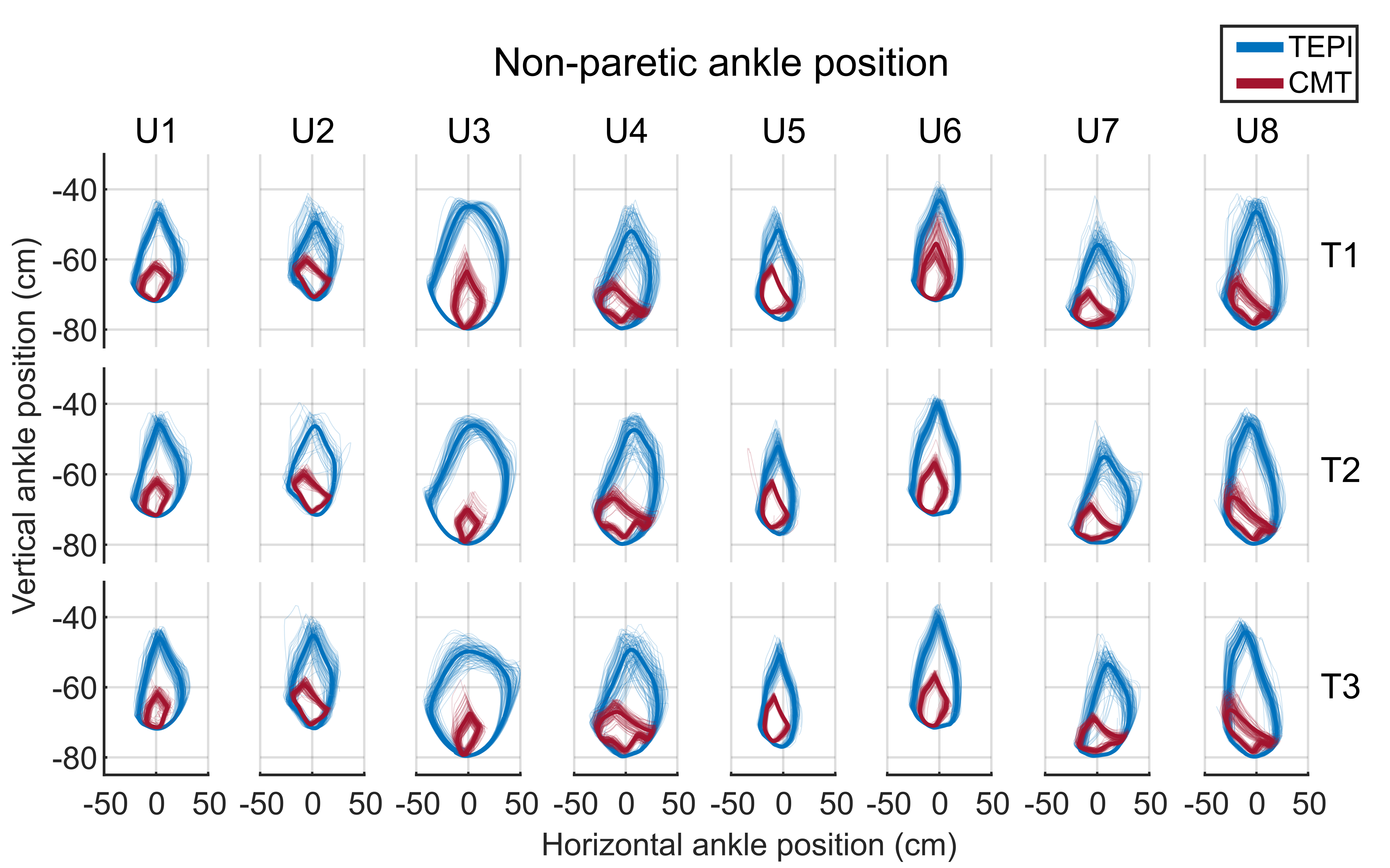}
\caption{\textbf{Non-paretic ankle trajectories during training.} For each patient (U1-U8), non-paretic limb ankle trajectories across training blocks (T1-T3) are displayed for the two training types: TEPI (blue) and CMT (red). Thin lines indicate individual strides while thick lines indicate the mean trajectory for a patient in a block of a training type.}
\label{fig:non-paretic-ankle}
\end{figure}

\begin{figure}
    \centering
    \includegraphics[width=1.0\columnwidth]{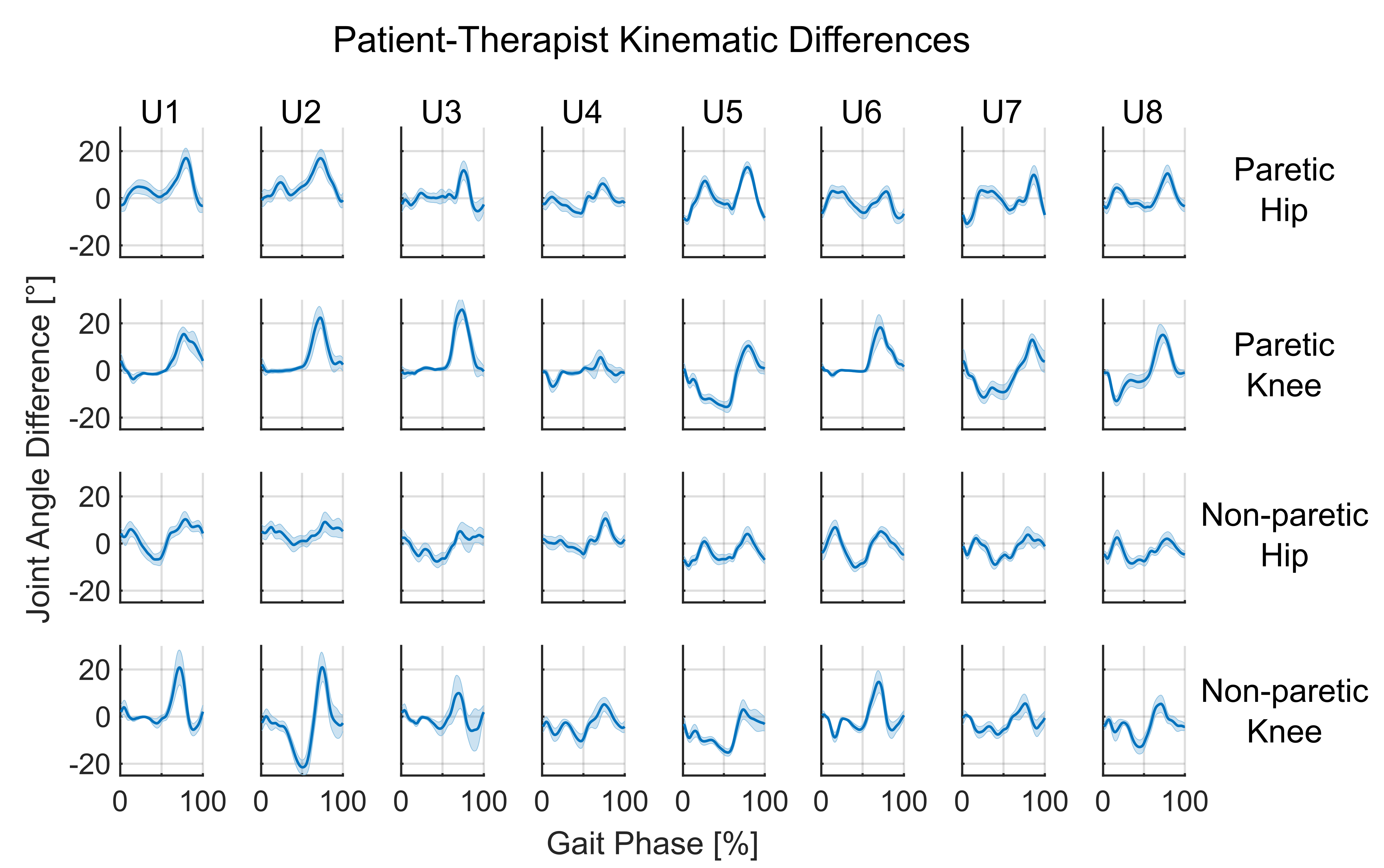}
    \caption{\textbf{Instantaneous differences between patient and therapist kinematics during TEPI training.} Differences between patient and therapist joint angles are plot across gait cycles for each user. Solid lines indicate mean time-series difference while shaded regions indicate $\pm$ one standard deviation across strides grouped for the 3 blocks of TEPI training.}
    \label{fig:sup:diff}
\end{figure}

\begin{figure}
    \centering
    \includegraphics[width=0.5\linewidth]{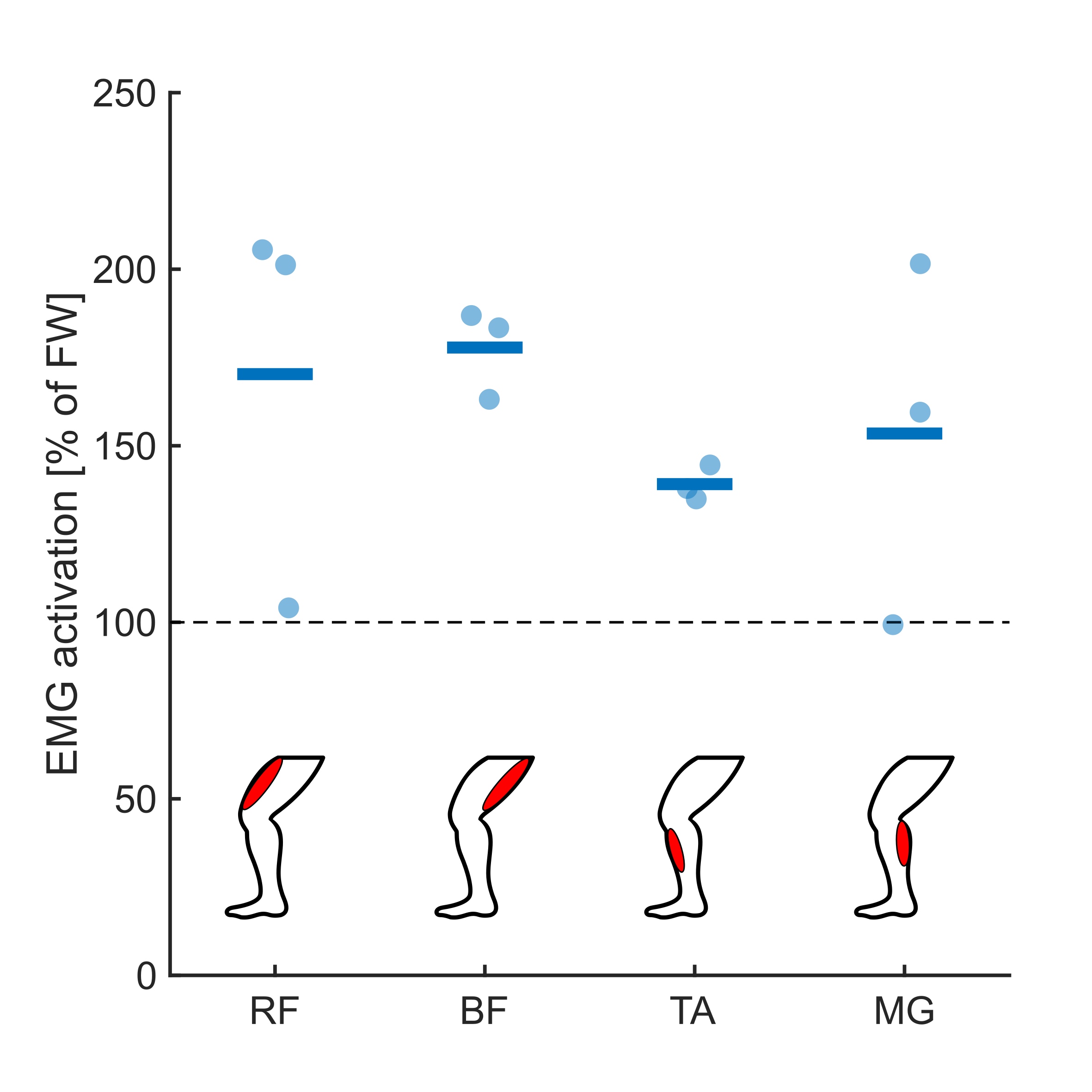}
    \caption{\textbf{Muscle activation during transparent exoskeleton walking in healthy pilot users.} Pilot EMG data was collected from healthy participants (N = 3) during treadmill walking at 0.3 m/s while wearing the exoskeleton in transparent mode (commanded interaction torque = 0 Nm). Participants walked for three minutes without the exoskeleton (free walking) and with the exoskeleton in transparent mode. Muscle activation is reported as transparent-mode EMG normalized with respect to the mean EMG of each participant’s free-walking (FW) trial. For each muscle, circular markers show the participant means and the horizontal bar indicates the group mean.}
    \label{fig:healthy_emg}
\end{figure}

\begin{table}
    \centering
    \begin{tabular}{ |p{2cm}||p{1.5cm}|p{1.5cm}|p{1.5cm}|p{1.5cm}|p{1.5cm}|p{1.5cm}|  }
     \hline
     Patient &\multicolumn{2}{|c|}{Training Block 1} &\multicolumn{2}{|c|}{Training Block 2} &\multicolumn{2}{|c|}{Training Block 3}   \\
     &\multicolumn{2}{|c|}{(10 mins)} &\multicolumn{2}{|c|}{(10 mins)} &\multicolumn{2}{|c|}{(10 mins)}   \\
     \hline
     & $K_p [\frac{\text{Nm}}{\text{rad}}]$ & $K_t [\frac{\text{Nm}}{\text{rad}}]$ & $K_p [\frac{\text{Nm}}{\text{rad}}]$ & $K_t [\frac{\text{Nm}}{\text{rad}}]$ & $K_p [\frac{\text{Nm}}{\text{rad}}]$ & $K_t [\frac{\text{Nm}}{\text{rad}}]$ \\
     \hline
     U1  & 49 & 49 & 49 &49 &49 &49 \\
     U2  & 49 & 49 & 49 &49 &- &- \\
     U3  & 49 & 49 & 58 & 49 &58 &49 \\
     U4  & 64 & 64 & 49 &49 &49 &49\\
     U5  & 49 & 49 & 49 &49 &49 &36 \\
     U6  & 49 & 49 & 49 &25 &49 &25 \\
     U7  & 60 & 60 & 60 &60 &60 &60 \\
     U8  & 60 & 60 & 60 &45 &60 &30 \\
     \hline
    \end{tabular}
    \caption{\textbf{Virtual stiffness combinations across training blocks for each patient.} The therapist dictated stiffness changes by requesting more or less feedback from the patient, adjusting $K_t$ accordingly, and by modulating haptic feedback transmitted to the patient through changes in $K_p$. Patient U2 performed only two of the three blocks due to a technical issue with the robotic system.}
    \label{tab:stiffness}
\end{table}

\begin{table}
\centering
\begin{tabular}{llllcc}
\hline
\textbf{Limb} & \textbf{Outcome} & \textbf{Effect} & \textbf{DoF} & \textbf{F} & \textbf{p} \\
\hline

Paretic & Work area & Block        & (2, 14) & 3.040  & 0.080 \\
        &           & \textbf{Type}        & (1, 7)  & \textbf{13.021} & \textbf{0.009} \\
        &           & Block $\times$ Type & (2, 14) & 2.870  & 0.090 \\[4pt]

Paretic & Step length & Block        & (2, 14) & 2.387  & 0.128 \\
        &             & \textbf{Type}        & (1, 7)  & \textbf{25.362} & \textbf{0.002} \\
        &             & Block $\times$ Type & (2, 14) & 0.203  & 0.819 \\[4pt]

Paretic & Step height & Block        & (2, 14) & 0.822  & 0.460 \\
        &             & \textbf{Type}        & (1, 7)  & \textbf{12.916} & \textbf{0.009} \\
        &             & Block $\times$ Type & (2, 14) & 0.508  & 0.613 \\[4pt]

Non-paretic & Work area & Block        & (2, 14) & 1.148  & 0.345 \\
            &           & \textbf{Type}        & (1, 7)  & \textbf{16.046} & \textbf{0.005} \\
            &           & Block $\times$ Type & (2, 14) & 0.680  & 0.523 \\[4pt]

Non-paretic & Step length & Block        & (2, 14) & 1.893  & 0.187 \\
            &             & \textbf{Type}        & (1, 7)  & \textbf{60.698} & \textbf{\textless{} .001} \\
            &             & Block $\times$ Type & (2, 14) & 0.166  & 0.849 \\[4pt]

Non-paretic & Step height & Block        & (2, 14) & 3.176  & 0.073 \\
            &             & \textbf{Type}        & (1, 7)  & \textbf{10.678} & \textbf{0.014} \\
            &             & Block $\times$ Type & (2, 14) & 0.558  & 0.585 \\

\hline
\end{tabular}
\caption{\textbf{Repeated measures ANOVA results for comparisons of spatial gait measures between TEPI and CMT.} Block indicates the effect of training block number (T1, T2, T3) and Type indicates the effect of training type (TEPI, CMT). Block x Type is the interaction between these effects.}
\label{tab:anova-spatial}
\end{table}

\begin{table}
\centering
\begin{tabular}{llllcc}
\hline
\textbf{Limb} & \textbf{Outcome} & \textbf{Effect} & \textbf{DoF} & \textbf{F} & \textbf{p} \\
\hline

Paretic & Rectus Femoris & Block        & (2, 8) & 0.682 & 0.533 \\
        &                & Type         & (1, 4) & 0.866 & 0.405 \\
        &                & Block $\times$ Type & (2, 8) & 0.743 & 0.506 \\[4pt]

Paretic & Biceps Femoris & Block        & (2, 8) & 0.608 & 0.568 \\
        &                & \textbf{Type}         & (1, 4) & \textbf{15.622} & \textbf{0.017} \\
        &                & Block $\times$ Type & (2, 8) & 0.388 & 0.691 \\[4pt]

Paretic & Tibialis Anterior & Block        & (2, 8) & 3.569 & 0.078 \\
        &                   & Type         & (1, 4) & 2.294 & 0.204 \\
        &                   & Block $\times$ Type & (2, 8) & 0.004 & 0.996 \\[4pt]

Paretic & Medial Gastrocnemius & Block        & (2, 8) & 1.782 & 0.229 \\
        &                      & \textbf{Type}         & (1, 4) & \textbf{10.727} & \textbf{0.031} \\
        &                      & Block $\times$ Type & (2, 8) & 0.150 & 0.863 \\

\hline
\end{tabular}
\caption{\textbf{Repeated measures ANOVA results for comparisons of muscle activation between TEPI and CMT.} Block indicates the effect of training block number (T1, T2, T3) and Type indicates the effect of training type (TEPI, CMT). Block x Type is the interaction between these effects.}
\label{tab:anova-muscle}
\end{table}

\begin{table}
\centering
\begin{tabular}{llllcc}
\hline
\textbf{User} & \textbf{Outcome} & \textbf{Effect} & \textbf{DoF} & \textbf{F} & \textbf{p} \\
\hline

Patient & Heart Rate & \textbf{Block} & (2, 12) & \textbf{7.805} & \textbf{0.007} \\
        &            & Type           & (1, 6)  & 0.413          & 0.544 \\
        &            & Block $\times$ Type & (2, 12) & 2.749      & 0.104 \\[4pt]

Patient & Perceived Exertion & \textbf{Block} & (2, 12) & \textbf{14.129} & \textbf{\textless{} .001} \\
        &                    & Type           & (1, 6)  & 0.068            & 0.803 \\
        &                    & Block $\times$ Type & (2, 12) & 0.054      & 0.948 \\

\hline
\end{tabular}
\caption{\textbf{Repeated measures ANOVA results for comparisons of heart rate and rating of perceived exertion between TEPI and CMT.} Block indicates the effect of training block number (T1, T2, T3) and Type indicates the effect of training type (TEPI, CMT). Block x Type is the interaction between these effects.}
\label{tab:anova-hr-rpe}
\end{table}

\end{document}